%% file: neurips_2022.tex
\documentclass{article}

\usepackage[preprint,nonatbib]{neurips_2022}

\usepackage{mathtools} 
\usepackage{booktabs} 
\usepackage{tikz} 
\usepackage{graphicx,multicol}
\usepackage{caption}
\usepackage{subcaption}
\usepackage{multirow}
\usepackage{comment}
\usepackage{float}
\usepackage{bbm}

\usepackage[utf8]{inputenc} 
\usepackage[T1]{fontenc}    
\usepackage{url}            
\usepackage{booktabs}       
\usepackage{amsfonts}       
\usepackage{nicefrac}       
\usepackage{microtype}      
\usepackage{xcolor}         
\usepackage[ruled,vlined,linesnumbered]{algorithm2e}
\usepackage{enumitem}

\newcommand{\nlpartynospace}{non-label party}
\newcommand{\nlparty}{\nlpartynospace~}
\newcommand{\lpartynospace}{label party}
\newcommand{\lparty}{\lpartynospace~}

\newcommand{\calX}{\mathcal{X}}

\newcommand{\Real}{\mathbb{R}}

\newtheorem{theorem}{Theorem}[section]
\newtheorem{lemma}[theorem]{Lemma}

\definecolor{Red}{rgb}{1,0,0}
\definecolor{Green}{rgb}{0,0.7,0}
\definecolor{Blue}{rgb}{0,0,1}
\definecolor{Red}{rgb}{0.6,0,0}
\definecolor{Orange}{rgb}{1,0.5,0}

\title{Label Leakage and Protection from Forward Embedding  in Vertical Federated Learning}

\author{%
  Jiankai Sun\thanks{Bytedance Inc. Corresponds to: \texttt{\{jiankai.sun, chong.wang\}@bytedance.com}} 
   \And
  Xin Yang
   \AND
  Yuanshun Yao
   \And
  Chong Wang
}

\begin{document}

\maketitle

\begin{abstract}
  \input{tex/abstract}
\end{abstract}

\input{tex/outline}

\clearpage

\bibliographystyle{plain}
\bibliography{reference}



\clearpage

\appendix
\input{tex/appendix}

\end{document}

%% file: tex/abstract.tex
Vertical federated learning (vFL) has gained much attention and been deployed to solve machine learning problems with data privacy concerns in recent years. However, some recent work demonstrated that vFL is vulnerable to privacy leakage even though only the forward intermediate embedding (rather than raw features) and backpropagated gradients (rather than raw labels) are communicated between the involved participants. As the raw labels often contain highly sensitive information, some recent work has been proposed to prevent the label leakage from the backpropagated gradients effectively in vFL. However, these work only identified and defended the threat of label leakage from the backpropagated gradients. None of these work has paid attention to the problem of label leakage from the intermediate embedding. In this paper, we propose a practical label inference method which can steal private labels effectively from the shared intermediate embedding even though some existing protection methods such as label differential privacy and gradients perturbation are applied. The effectiveness of the  label attack is inseparable from the correlation between the intermediate embedding and corresponding private labels. To mitigate the issue of label leakage from the forward embedding, we add an additional optimization goal at the \lparty to limit the label stealing ability of the adversary by minimizing the distance correlation between the intermediate embedding and corresponding private labels. We conducted massive experiments to demonstrate the effectiveness of our proposed protection methods. 

%% file: tex/outline.tex
\input{tex/intro}
\input{tex/relatedwork}
\input{tex/preliminaries}
\input{tex/methods}
\input{tex/experiments}

\input{tex/conclusion}

%% file: tex/intro.tex
\section{Introduction}
\label{sec:introduction}








With increasing concerns over data privacy in machine learning, \textit{Split learning}~\cite{gupta2018distributed,vepakomma2018split} and \textit{vertical federated learning} (vFL)~\cite{yang2019federated} have gained attention and been deployed to solve practical problems such as online advertisements conversion prediction tasks ~\cite{li2021label,sunDefending2021}. Both techniques can jointly learn a machine learning model by splitting the execution of the corresponding deep neural network on a layer-wise basis  without raw data sharing. 

\begin{figure*}[ht!]
  \centering
  \includegraphics[width=0.85\linewidth]{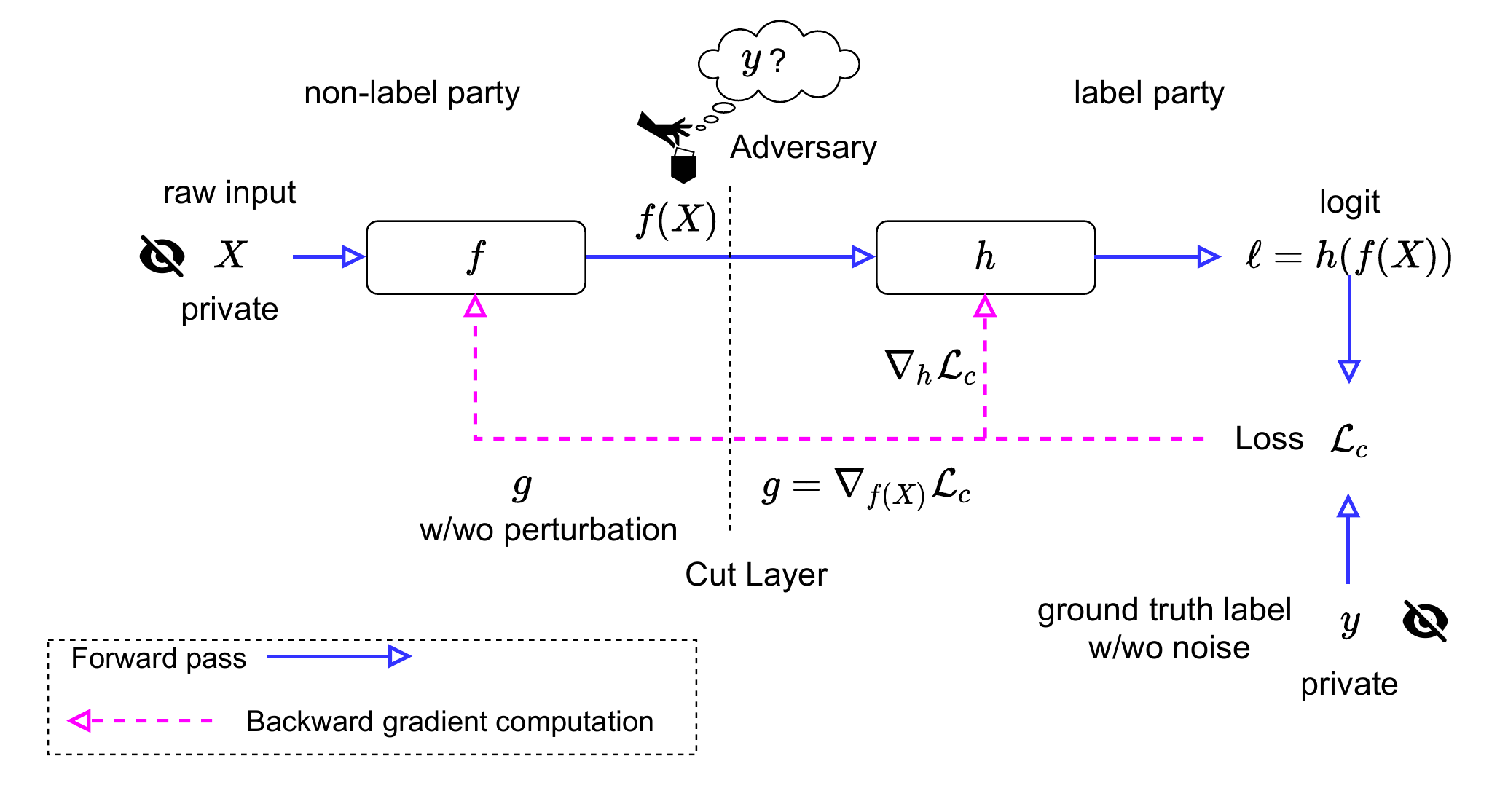}
  \caption{Attack scenario of label leakage in two-party split learning. During training, the  \nlparty sends the cut  layer embedding $f(X)$ to the \lpartynospace, and the \lparty completes the rest of the forward computation to obtain the training loss $\mathcal{L}_c$. To compute the gradients w.r.t model parameters, the \lparty starts backpropagation by computing the gradients $\nabla_h \mathcal{L}_c$. To allow the \nlparty to also compute gradients of its parameters, the \lparty also computes the gradients w.r.t the cut layer embedding $g = \nabla_{f(X)} \mathcal{L}_c$ and sends it to the \nlpartynospace. In this scenario, the goal of the attacker is to infer the label from the forward cut layer embedding $f(X)$.   }
 \label{fig:split-learning-setup} 
 \end{figure*}

The detailed training process of vFL (including forward pass and backward gradients computation) can be seen in Figure~\ref{fig:split-learning-setup}. During the forward pass, the party without labels (\textit{\nlpartynospace}) sends the intermediate layer (\textit{cut layer}) outputs rather than the raw data to the party with labels (\textit{\lpartynospace}), and the \lparty completes the rest of the forward computation to obtain the training loss. To compute the gradients with respect to model parameters in the backward gradients computation phase, the \lparty initiates backpropagation from its training loss and computes its own parameters' gradients. The \lparty also computes {the gradients with respect to the cut layer outputs} and sends this information back to the \nlpartynospace, so that the \nlparty can use the chain rule to  compute gradients of its  parameters. 

Even though only the intermediate computations of the cut layer embedding (rather than raw features) and backpropagated gradients (rather than raw labels) are communicated between the two parties,  existing work \cite{li2021label,sunDefending2021} demonstrated vFL is vulnerable to privacy leakage. For example, \cite{li2021label} demonstrated that an adversarial \nlparty can leverage both norm and direction of the backpropagated gradients to infer the private class labels accurately (AUC is almost $1.0$). 
As the raw labels often contain highly sensitive information (\textit{e.g.}, what a user has purchased (in online advertising) or whether a user has a disease or not (in disease prediction)~\cite{vepakomma2018split}, understanding of the threat of label leakage and its protection in vFL is particularly important.

Some recent work \cite{li2021label,labeldp} have been proposed to prevent the label leakage. For example, \cite{labeldp} leveraged randomized responses to flip labels and use the generated noisy version labels to compute the loss functions. They proved that their proposed algorithms can achieve label differential privacy (DP) \cite{DPDwork2006}. \cite{li2021label} proposed to add optimized Gaussian noise to perturbate the backpropagated gradients so that positive and negative gradients cannot be distinguished after perturbation. The optimized amount of added noise is calculated by minimizing the sum KL divergence between the perturbed gradients. Both methods are effective on preventing the label leakage  from the backpropagated gradients in the setting of vFL.

However, besides backpropagated gradients, private labels can also be inferred by adversarial parties from the forward cut layer embedding sent from the \nlparty to the \lpartynospace. With the model training, the cut layer embedding can be learned to have a correlation with private labels and hence be used to distinguish different labels by adversarial parties. For example, adversarial \nlparty parties can do clustering on these cut layer embeddings and then assign labels to clusters based on their sizes. For example, in the imbalanced binary classification settings such as online advertising and disease prediction, the larger cluster can be assigned negative labels and the smaller one will get positive labels. Particularly, in this paper, we develop a practical label inference method based on the technique of spectral attack ~\cite{tlm18} to predict the private labels. Our inference method can steal the private labels from the cut layer embedding effectively even protection methods proposed by \cite{li2021label} and \cite{labeldp} are applied (since both methods are proposed to prevent the label leakage from backpropagated gradients). The effectiveness of  our label attack is inseparable from the correlation between the intermediate embedding and corresponding private labels. To mitigate the issue of label leakage from forward embedding, we propose to reduce the learned correlation between the cut layer embedding and private labels. We achieve this goal by letting the \lparty optimize an additional target which minimizes the distance correlation \cite{dcor2007} between the cut layer embedding and corresponding private labels. So that the adversarial party cannot distinguish positive and negative labels based on the limited correlation between the cut layer embedding and private labels.  



We summarize our contributions as: 1)
We propose a practical label leakage attack from the forward embedding in two-party split learning; 2) We propose a corresponding defense that minimizes the distance correlation between cut layer embedding and private labels. 3) We experimentally verify the effectiveness of our attack and defense. 4) Our work is the first we are aware of to identify and defend against the threat of label leakage from the forward cut layer embedding in vFL.

%% file: tex/relatedwork.tex

\section{Related Work}
\label{sec:related-work}

\textbf{Federated Learning.} 
\textit{Federated learning} (FL)~\cite{mcmahan2017communication} can be mainly classified into three categories: \textit{horizontal FL},  \textit{vertical FL}, and \textit{federated transfer learning} ~\cite{yang2019federated}. We focus on vertical FL (vFL) which partitions data by features (including labels). When the jointly trained model is a neural network, the corresponding setting is the same as split learning such as SplitNN~\cite{vepakomma2018split}. In the following, we use the term vFL and split learning interchangeably.


\textbf{Information Leakage in Split Learning.} Recently, researchers have shown that in split learning, even though the raw data (feature and label) is not shared,
sensitive information can still be leaked from the gradients and intermediate embeddings communicated between parties. For example, \cite{vepakomma2019reducing} and \cite{sunDefending2021} showed that \nlpartynospace's raw features can be leaked from the forward cut layer embedding. We differ from them by studying the leakage of labels rather than raw features. In addition, \cite{li2021label} studied the label leakage problem but the leakage source was the backward gradients rather than forward embeddings. To the best of our knowledge, there is no prior work studying the label leakage problem from the forward embeddings.

\textbf{Information Protection in vFL.} 
There are three main categories of information protection techniques in vFL:
\textbf{1)} cryptographic methods such as {secure multi-party computation} \cite{bonawitz2017practical}; \textbf{2)} system-based methods including {trusted execution environments} \cite{subramanyan2017formal}; and \textbf{3)} perturbation methods that add noise to the communicated messages~\cite{abadi2016deep, mcmahan2017learning,erlingsson2019amplification,cheu2019distributed,zhu2019dlg}. 
Our protection belongs to the third category. In this line of work, \cite{labeldp} generated noisy labels by randomized responses to achieve label differential privacy guarantees. \cite{li2021label} added optimized Gaussian noise to perturb the backpropagated gradients to confuse positive and negative gradients.

%% file: tex/preliminaries.tex
\section{Preliminaries}
\label{sec:preliminaries}

We introduce the background of vFL framework.  vFL \cite{yang2019federated} splits a deep neural network model by layers during the training and inference time.

The training phase of vFL includes two steps as shown in Figure~\ref{fig:split-learning-setup}: forward pass and backward gradient computation.

 \textbf{Forward Pass.} We consider two parties that jointly train a composite model $h \circ f$ on a binary classification problem over the domain $\calX \times \{0, 1\}$. The \nlparty owns the representation-generating function $f: \calX \to \mathbb{R}^d$ and each example's raw feature $X \in \calX$, while the \lparty owns the logit-predicting function $h: \mathbb{R}^d \to \mathbb{R}$ and each example's label $y \in \{0, 1\}$\footnote{
For simplicity, we consider the scenario where \lparty owns only labels without any additional features. Our attack and defense can be applied to other vFL scenarios without major modifications.
}. The raw input $X$ and label $y$ are considered as private information by the \nlparty and \lparty respectively and should be kept to themselves only.
Let $\ell = h(f(X))$ be the logit of the positive class whose predicted probability is given through the sigmoid function: $\widetilde{p}_1 = 1/(1 + \exp(-\ell))$. We measure the loss of such prediction through the binary cross entropy loss $\mathcal{L}_c$. 
When computing $\mathcal{L}_c$, since the ground-truth label $y$ is sensitive information, the \lparty can provide noisy labels instead of real labels to prevent label leakage. For example, \cite{labeldp} relied on a randomized response algorithm to flip labels and provided label DP guarantee for these generated flipped labels. In this phrase, $f(X)$ is the forward embeddings and the label leakage source that we consider.

\textbf{Backward Gradient Computation}. To perform gradient descent, the \lparty first computes the gradient of the loss $\mathcal{L}_c$ w.r.t the logit $\nabla_{\ell}\mathcal{L}_c$. Using the chain rule, the \lparty then computes the gradient of $\mathcal{L}_c$ w.r.t its function $h$'s parameters and update the gradients. To allow the \nlparty to learn $f$, the \lparty needs to additionally compute the gradients w.r.t cut layer embeddings $f(X)$ and communicate it to the \nlpartynospace. We denote this gradient by $g \coloneqq \nabla_{f(X)} \mathcal{L}_c $. Prior work has demonstrated that the vanilla $g$ can leak the label information. For example, \cite{li2021label} showed that adversarial parties can leverage either direction or norm of $g$ to recover labels. They proposed a random perturbation technique called Marvell which we will consider as a baseline to compare with. After receiving $g$ (w/wo perturbation) from the \lpartynospace, the \nlparty continues the backpropagation on $f$'s parameters using chain rule and updates the corresponding gradients. 


During model inference, the \nlparty computes $f(X)$ and sends it to the \lparty who then executes the rest of forward computation to compute the prediction probability. 

%% file: tex/methods.tex
\input{tex/threat_model}

\input{tex/spectral_attack}

\section{Defense against Label Leakage Attack}
\label{sec:label_protection}

In this section, we first talk about the protection objectives and then show the details of our practical protection method on defending the label leakage from the forward embedding.

\subsection{Label Protection Objectives}
\label{sec:label_protection_objectives}

The defender, which is the \lparty who is the victim of the attacker, aims to protect the label information during the procedure while maintaining the model's utility. 
It has to achieve reasonable trade-offs between two objectives: privacy and utility objective.

\textit{Privacy Objective:} the leak AUC of the recovered labels that the adversarial party estimate from the cut layer embedding should be close to $0.5$ (like a random guess).

\begin{equation}
    \min  \mathop{\mathbb{E}}_{Y \sim \mathcal{D}_{label}} [|AUC(Y, Y') - 0.5|]
\end{equation}

where $Y'$ is the estimated label by an attacker and $Y$ is the corresponding ground-truth. 

\textit{Utility Objective:} The composition model $h \circ f$ jointly trained
with split learning needs to have high classification performance (evaluated by AUC) on an unseen test set.

\begin{equation}
    \max  \mathop{\mathbb{E}}_{(X, Y)\sim \mathcal{D}_{test}} [AUC(Y, \mathbbm{1}{(\widetilde{p}_1 \geq \lambda})))]
\end{equation}

where $\widetilde{p}_1 = 1/(1 + \exp(-h(f(X))))$ is the predicted probability of being positive for $X$ and $\mathbbm{1}$ is an indicator function with a tuning parameter $\lambda$.


\subsection{Defense Details}
\label{subsec:dcor}

 With the model training, the function $f$ gains some high-level representation of raw input $X$ and learns the correlation between $f(X)$ and $Y$. $f(X)$ can be learned as an indicator of private labels. Hence spectral attack can leverage $f(X)$ to steal private labels. One strategy to prevent  label leakage is that the \lparty asks the \nlparty to send the raw input $X$ directly. However, this strategy violates the protocol of vFL and \nlparty has risks of revealing their raw input to the other party. Another strategy is that the \lparty forces \nlparty to limit its power of $f$ so that $f(X)$ cannot be a good indicator of private labels. However, this strategy is not practical since it violates the setting of the adversary's capability defined in our threat model (see Section ~\ref{sec:threatmodel})and may also break the utility objective as discussed in Section~\ref{sec:label_protection_objectives}. In this paper, to prevent the label leakage from the forward embedding, we would like to let the forward embeddings not be a good proxy of the private labels while meeting the requirements of our threat model (Section ~\ref{sec:threatmodel}) and label protection objectives (Section ~\ref{sec:label_protection_objectives}). Particularly, we leverage distance correlation to achieve this goal. 

Distance correlation can be used to measure the statistical dependence between two paired random vectors with arbitrary dimensions (not necessarily equal) ~\cite{dcor2007}. It can measure both linear and nonlinear associations between two vectors~\cite{nopeek2018}. Another advantage is it can be optimized and easily fit into the conventional optimization framework (e.g. minimizing the cross entropy loss).

Given $f(X)$ and $Y$ from a mini-batch, we can use $dCor(f(X), Y)$ where $f(X) \in  \Real^{n \times d}$ and $Y \in \Real^{n \times 1}$ to measure the dependence between $f(X)$ and $Y$, even though  $f(X)$ and $Y$ share different dimensions. The detailed calculation of $dCor(f(X), Y)$ can be shown as follows:

\begin{enumerate}[leftmargin=*]
\setlength\itemsep{0pt}
    \item Compute the $n \times n$ distance matrices $\mathbf{s}$ and $\mathbf{t}$ containing all pairwise distances for each  sample pair. Each element  $\mathbf{s}_{(j,k)}$ in $\mathbf{s}$ is computed as: $\mathbf{s}_{(j,k)}=\|f(X)_{j}-f(X)_{k}\|$ and each element $\mathbf{t}_{(j,k)}$ in $\mathbf{t}$ is $\mathbf{t}_{(j,k)}=\|Y_{j}-Y_{k}\|$ where $j,k=1,2,\ldots ,n$, $f(X)_j$ represents the the $j$-th row of $f(X)$, $Y_j$ denotes the label ($0$ or $1$) of the $j$-th sample, and $\| \cdot \|$ denotes Euclidean norm.
    \item  Normalize $\mathbf{s}$ and $\mathbf{t}$ by taking all doubly centered distances to get $\mathbf{A}$ and $\mathbf{B}$ respectively. Particularly, $\mathbf{A}_{(j,k)} = \mathbf{s}_{(j,k)} - \overline{\mathbf{s}}_{j,\cdot} - \overline{\mathbf{s}}_{\cdot,k} + \overline{\mathbf{s}}_{\cdot,\cdot}$ and $\mathbf{B}_{(j,k)} = \mathbf{t}_{(j,k)} - \overline{\mathbf{t}}_{j,\cdot} - \overline{\mathbf{t}}_{\cdot,k} + \overline{\mathbf{t}}_{\cdot,\cdot}$, where $\overline{\mathbf{s}}_{j,\cdot}$ and $\overline{\mathbf{t}}_{j,\cdot}$  are the $j$-th row mean of  $\mathbf{s}$ and $\mathbf{t}$ respectively, $\overline{\mathbf{s}}_{\cdot,k}$ and $\overline{\mathbf{t}}_{\cdot, k}$  are the $k$-th column mean of  $\mathbf{s}$ and $\mathbf{t}$ respectively, and  $\overline{\mathbf{s}}_{\cdot,\cdot}$ and $\overline{\mathbf{t}}_{\cdot, \cdot}$  are grand mean  $\mathbf{s}$ and $\mathbf{t}$ respectively.
    \item Calculate the distance covariance $dCov^2(f(X),Y)$  as the mean of the dot product of $\mathbf{A}$ and $\mathbf{B}$. Particularly, $dCov^2(f(X),Y) = \frac{1}{n^2} \sum_{j=1}^{n} \sum_{k=1}^{n} \mathbf{A}_{(j,k)} \mathbf{B}_{(j,k)}$. Similarly, compute $dCov^2(f(X),f(X))$ and $dCov^2(Y,Y)$.
    \item Compute the distance correlation $dCor(f(X), Y)$ as $ \frac{dCov^2(f(X),Y)}{\sqrt{dCov^2(f(X),f(X)) \times dCov^2(Y,Y)}}$

\end{enumerate}

The distance correlation $dCor(f(X), Y)$ is zero if and only if the $f(X)$ and $Y$ are independent, and then the adversary has no way to infer $Y$ from $f(X)$ under this scenario. The intuition of our protection method is to make label $Y$ and embedding $f(X)$ less dependent, and therefore reduces the likelihood of gaining information of $Y$ from $f(X)$. Particularly, we add an additional optimization target at the side of \lparty, and we name it as the distance correlation loss $\mathcal{L}_d$ which is calculated as follows:

\begin{equation}
\mathcal{L}_d =  \mathop{\mathbb{E}}_{(X,Y) \sim \mathcal{D}} (\log{(dCor(f(X), Y))}  
\end{equation}
 
 By minimizing the log of distance correlation \footnote{Empirically we find that adding $log(.)$ stabilizes training.}  (loss $\mathcal{L}_d$) during the model training, we achieve the goal of not letting $f(X)$ be a good proxy of $Y$.


Note that dCor computes pairwise distance between samples and requires $O(n^2)$ time complexity where $n$ is the batch size. In practice, there are some faster estimators of dCor~\cite{fastdcor2019,huang2017statistically}. In addition, dCor is sensitive to the $n$ and a larger $n$ can give a more accurate estimation of the distance correlation. 



\textbf{Combine with Training.}
The overall loss function computed at the \lparty side is a combination of two losses: distance correlation loss ($\mathcal{L}_d$) and normal label prediction loss ($\mathcal{L}_c$). In this paper, we focus on binary classification and use categorical cross-entropy as label prediction loss $\mathcal{L}_c$. Optimizing $\mathcal{L}_c$ makes sure the model maintain a good utility while optimizing $\mathcal{L}_d$ increases model privacy. Putting them in one framework can help us defend against the spectral attack while maintaining the accuracy of the primary learning task. The overall loss function is:

\begin{equation}
\mathcal{L} = \mathcal{L}_c  + \alpha_d \mathcal{L}_d 
\end{equation}

where $\alpha_d \geq 0$ is the control parameter for distance correlation. It can balance the trade-off between  utility and privacy. If we set $\alpha_d$ a large number, the cut layer embedding and the corresponding label will be more independent. Then the \lparty has great difficulty learning the model well and hence we cannot achieve our utility objective. However, if set $\alpha_d=0$, we fail on achieving the privacy objective but can get the best model utility. A good value of $\alpha_d$ can help us balance utility and privacy objectives. It's worth mentioning that only the \lparty deals with the optimization of $\mathcal{L}$ (unseen for the \nlpartynospace) while there is no change on the \nlpartynospace's side.

\textbf{Remarks.} We choose the two-party setting with binary classification because it is currently the typical scenario used in the vFL literature (i.e. \cite{lzw20,li2021label}) due to its popularity in the industry (online advertising, disease prediction etc). 
Our protection method can be easily extended to other  scenarios as we discussed in Section ~\ref{sec:extensions_of_protection_methods} of Appendix.

%% file: tex/threat_model.tex

\section{Threat Model}
\label{sec:threatmodel}

\textbf{Attacker's Goal.} During training, the attacker, which is the \nlparty, attempts to infer the private labels that belong to the \lpartynospace. Specifically, attacker wants to reconstruct the \lpartynospace's hidden label $y$ based on the intermediate embedding $f(X)$ for each training example. 


\textbf{Attacker's Capability.} We consider an \textit{honest-but-curious} \nlparty which follows the agreed training procedure but it actively tries to infer the label $y$ from $f(X)$ while not violating the procedure. The attacker cannot tamper with training by selecting which examples to include in a batch or sending incorrect cut layer embedding $f(X)$. 
The attacker can choose its own desired model architecture for the prediction function $f(.)$ as long as the cut layer embedding dimension matches with the input of $h(.)$. For example, the attacker has the freedom of selecting the depth and width of its neural network to increase the power of $f(X)$ on label guessing. 
Inferring label $y$ can be viewed as a binary classification problem where the (input, output) distribution is the induced distribution of $(f(X), y)$. The attacker can use any binary classifier $q: \Real^d \rightarrow \{0, 1\}$ to infer the labels. 

\textbf{Attacker's Auxiliary Information.} 
We allow the \nlparty to have population-level side information (i.e. ratio of the positive instances) specifically regarding the properties of (and the distinction between) the positive and negative class's cut-layer embedding distributions. The attacker may leverage such population-level side information to infer the private labels more accurately. 
However, we assume the \nlparty has \textit{no example-level side information} that is different example by example.

%% file: tex/spectral_attack.tex
\section{Label Leakage Attack}
\label{sec:label_leakage_problem}

We leverage spectral attack ~\cite{tlm18} to predict labels for corresponding cut layer embeddings. Spectral attack is a singular value decomposition (SVD) based outlier detection method. It shows that: 

\begin{lemma}[Lemma  in \cite{tlm18}]\label{lem:spectral_attack}
Let $D$, $W$ be two distributions over $\mathbb{R}^d$ with mean $\mu_D,\mu_W$ and covariance matrices $\Sigma_D,\Sigma_W$.
Let $F$ be a mixture distribution given by $F=(1-\epsilon)D+\epsilon W$ where $0<\epsilon<\frac{1}{2}$.
If $\|\mu_D-\mu_W\|_2^2\geq \frac{6\sigma^2}{\epsilon}$,
then the following statement holds:

\begin{equation}
    \Pr_{X\sim D}[|\langle X-\mu_F,v\rangle |>t] <  ~\epsilon,\\
    \Pr_{X\sim W}[|\langle X-\mu_F,v\rangle |<t]<  ~\epsilon.
\end{equation}

where $\mu_F$ is the mean of $F$ and $v$ is the top singular vector of the covariance matrix of $F$, and 
 $t>0$.
\end{lemma}


 Lemma \ref{lem:spectral_attack} motivates us to use the spectral attack to differentiate the embedding distribution between the positive samples ($\mathcal{F}^+$) and the negative samples ($\mathcal{F}^-$). If the mean of $\mathcal{F}^+$ and $\mathcal{F}^-$ are far away from each other, then we can use $|\langle f(X)-\mu_F,v\rangle |$ as the indicator to distinguish $\mathcal{F}^+$ and $\mathcal{F}^-$. Our attack works as following: 

\begin{enumerate}[leftmargin=*]
\setlength\itemsep{0pt}
    \item For each mini-batch of $f(X)$, compute its empirical mean $\mu_F$ and the top singular vector $v$ of  the corresponding covariance matrix. Then compute score $s = |\langle f(X)-\mu_F,v\rangle |$ for each sample in the mini-batch.
    \item Divide all samples into two clusters using score $s$ as the distance metric. We leverage some heuristic with population-level knowledge to determine which cluster belongs to which label. For example, if the dataset (\textit{i.e.} Criteo) is imbalanced and dominated by negative samples, then we assign negative labels to the larger cluster and positive labels to the smaller one. Note that when the dataset is balanced (i.e. positive ratio is $0.5$), it is true that the attacker cannot directly know which label to assign to which cluster. However, the attacker can employ a simple heuristic: assigning positive labels to the cluster with larger scores. 
\end{enumerate}

The detailed description of  our attack method can be seen in Algorithm ~\ref{alg:spectral_attck} in  Appendix ~\ref{sec:spectral_attack_alg}. We also talk about how to extend our attack method to multi-class settings in Appendix ~\ref{sec:extend_attack_to_multi_class}.

\section{Attack Evaluation}
\label{sec:attack_evaluation}

\begin{figure*}[h!]
\captionsetup[subfigure]{labelformat=empty}
  \begin{subfigure}{0.323\linewidth}
  \centering
    \includegraphics[width=\linewidth]{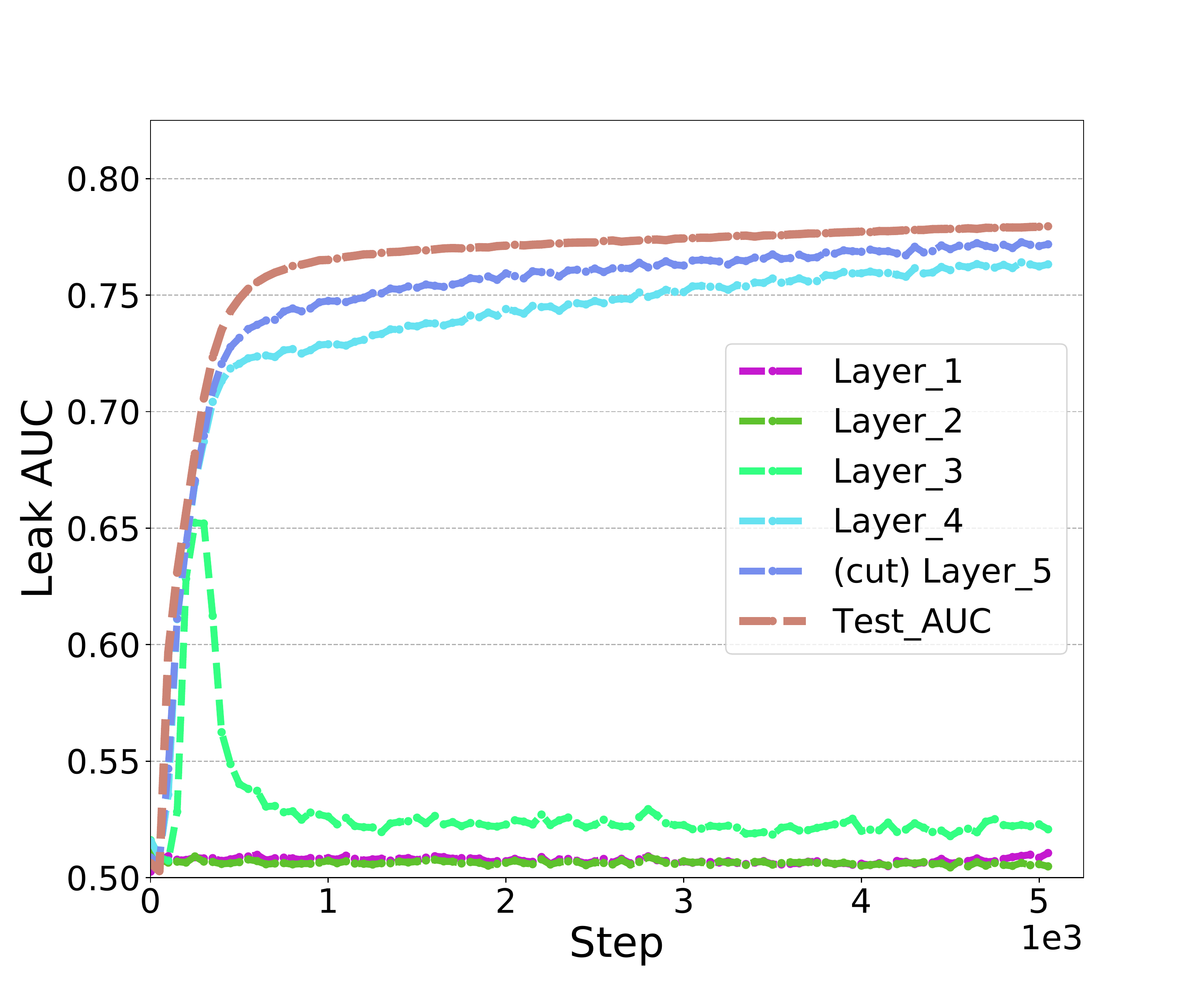}
      \caption{(a): Vanilla (w/o protection)}
  \end{subfigure}
  \begin{subfigure}{0.323\linewidth}
  \centering
    \includegraphics[width=\linewidth]{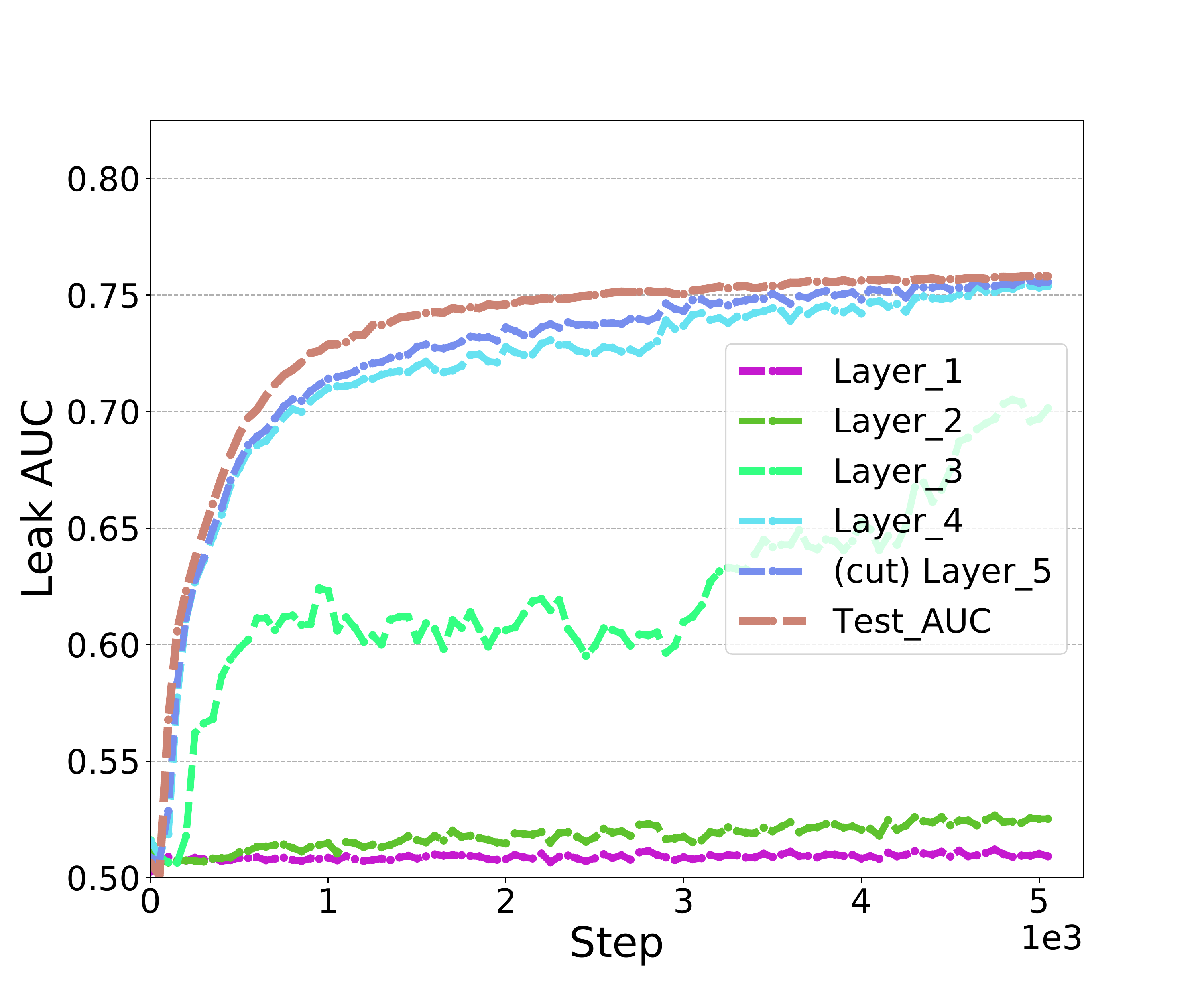}
    \caption{(b):   Marvell ($s=8.0$)}
  \end{subfigure}
    \begin{subfigure}{0.323\linewidth}
  \centering
    \includegraphics[width=\linewidth]{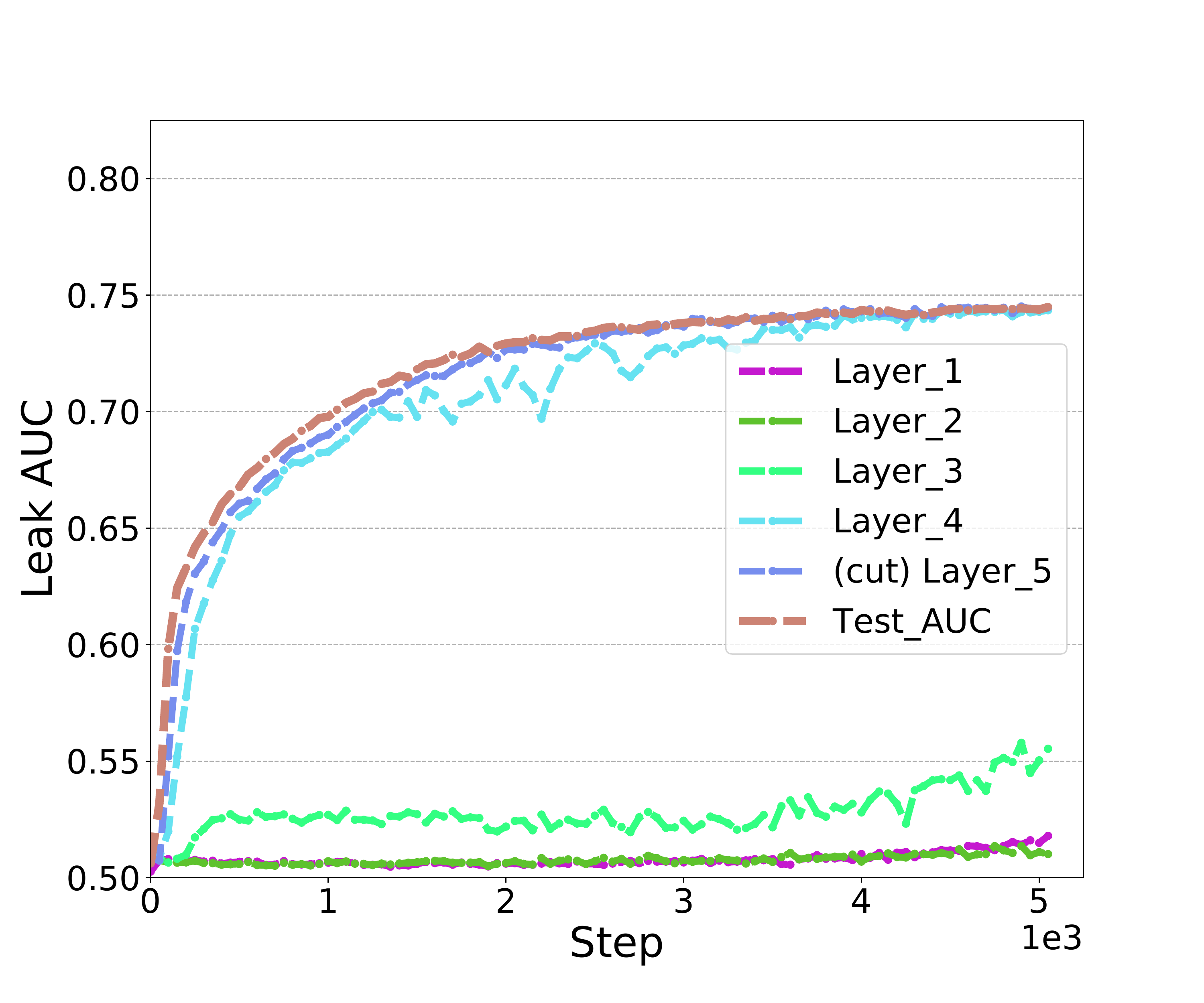}
      \caption{(c): Label DP ($\epsilon = 0.5$)}
  \end{subfigure}
   \caption{Effectiveness of Spectral Attack. (a): Spectral attack AUC of different layers without any protection; (b):Spectral attack AUC of different layers with Marvell~\cite{li2021label} ($s=8.0$) protection;  (c): Spectral attack AUC of different layers with Label DP ~\cite{labeldp} guarantee ($\epsilon = 0.5$).
}
 \label{fig:effectiveness_of_spectral_attack} 
 \end{figure*}

 \begin{figure*}[h!]
\captionsetup[subfigure]{labelformat=empty}
  \begin{subfigure}{0.245\linewidth}
  \centering
    \includegraphics[width=\linewidth]{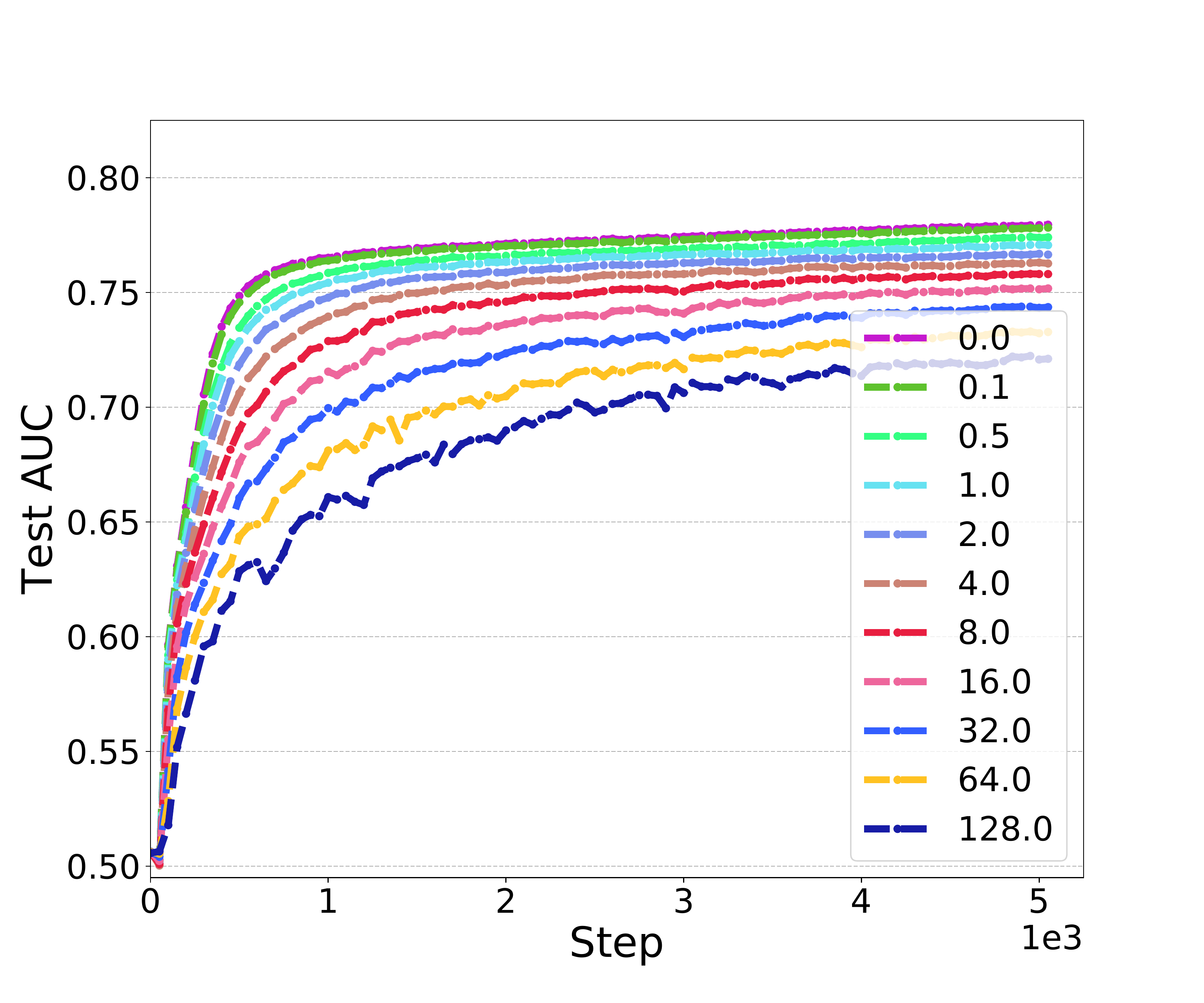}
    \caption{(a): Test AUC, Marvell}
  \end{subfigure}
 \begin{subfigure}{0.245\linewidth}
  \centering
    \includegraphics[width=\linewidth]{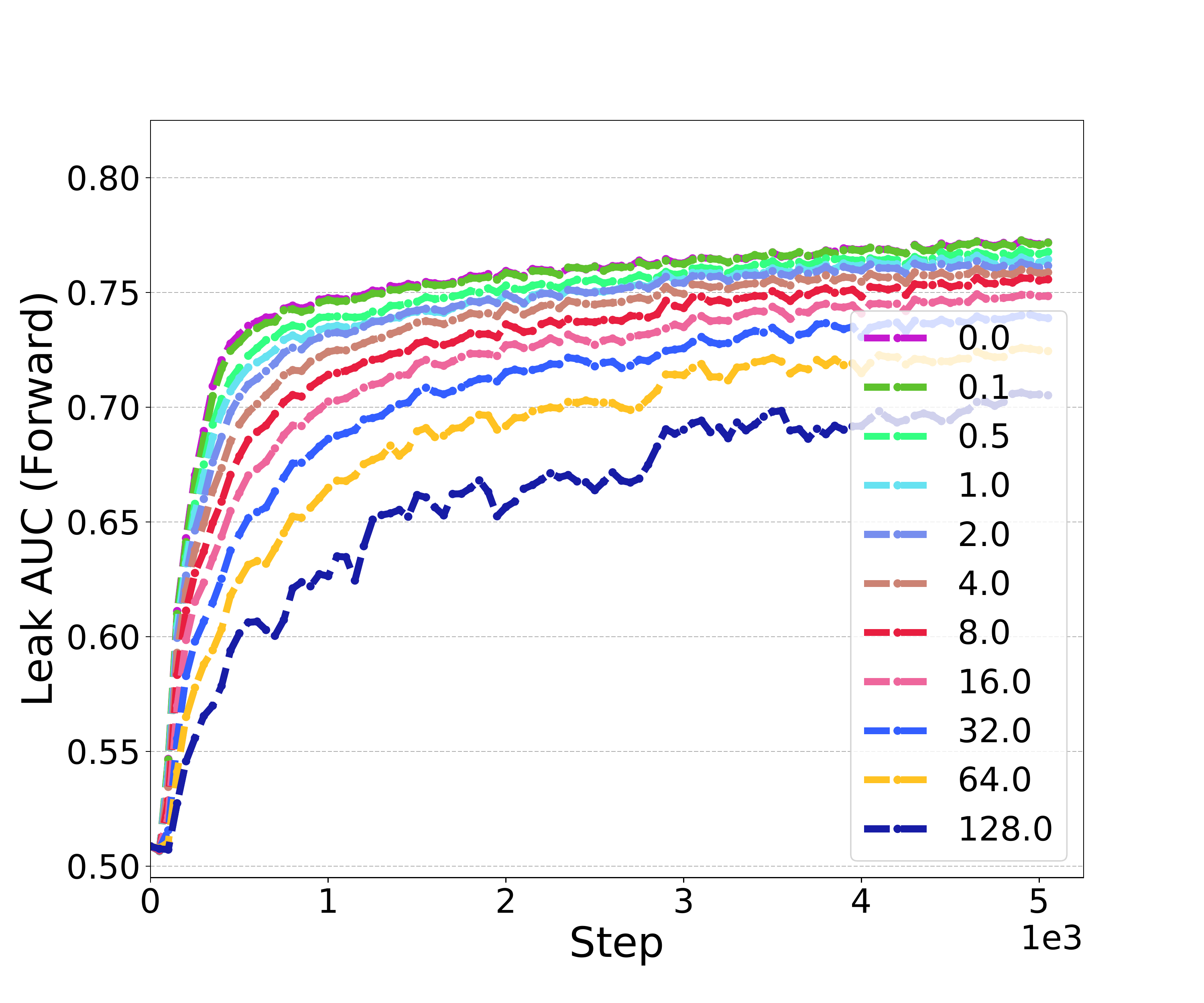}
      \caption{(b): Leak AUC, Marvell }
  \end{subfigure}
  \begin{subfigure}{0.245\linewidth}
  \centering
    \includegraphics[width=\linewidth]{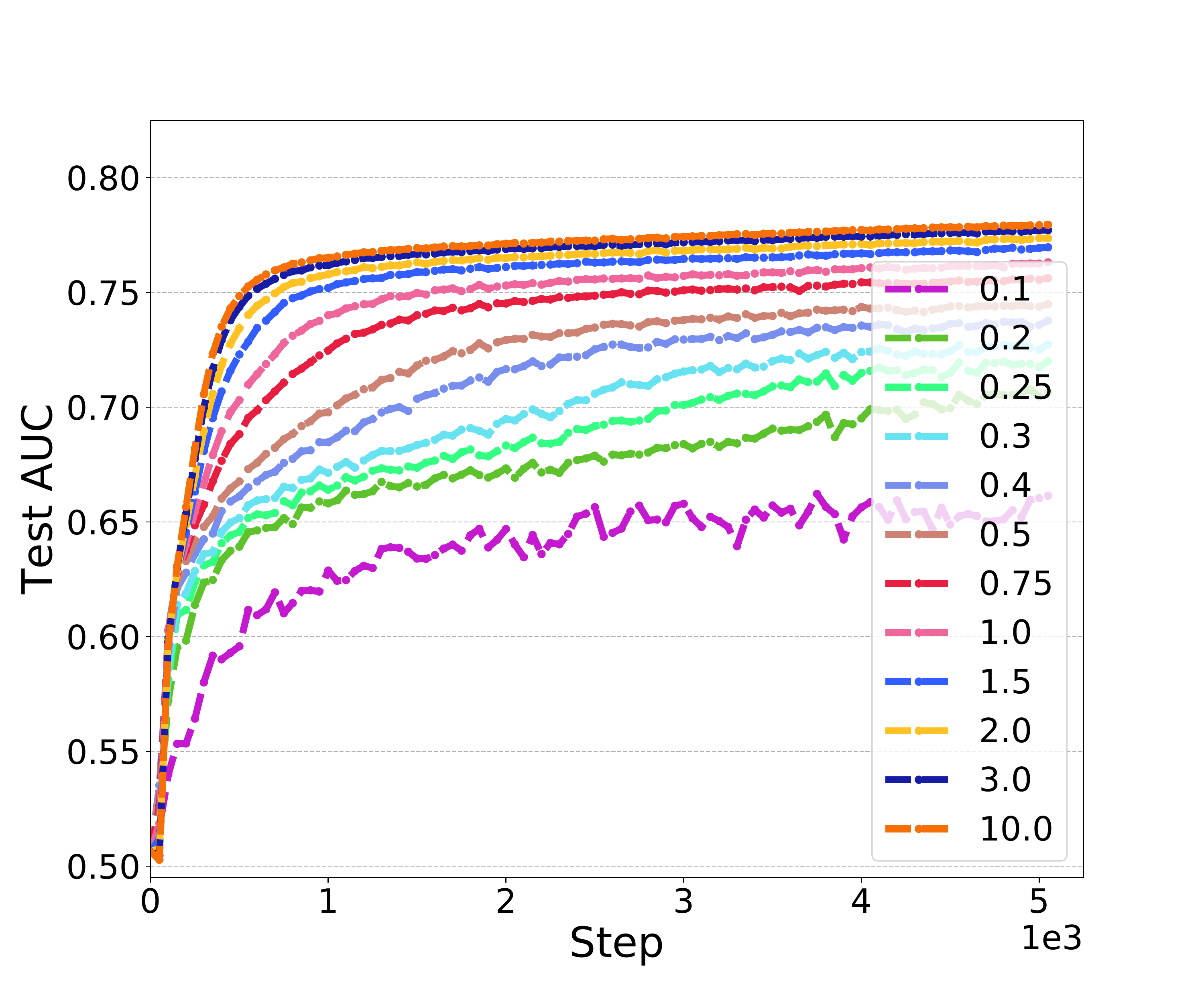}
    \caption{(c): Test AUC, Label DP}
  \end{subfigure}
  \begin{subfigure}{0.245\linewidth}
  \centering
    \includegraphics[width=\linewidth]{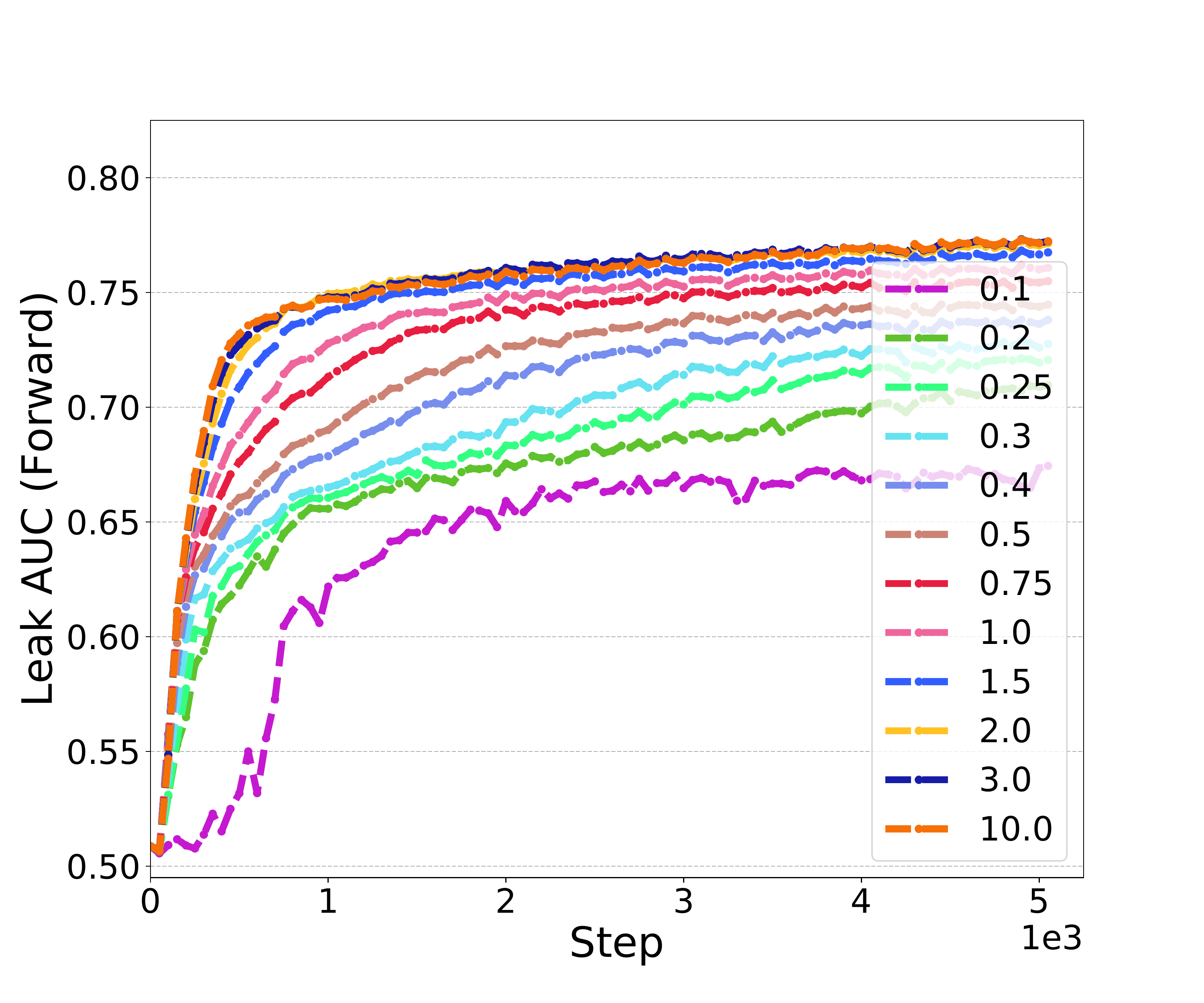}
      \caption{(d): Leak AUC, Label DP }
  \end{subfigure}
   \caption{Spectral Attack (leak) AUC is bounded by the model utility (test AUC). Figure (a) and (b) shows the test AUC and leak AUC on the cut layer protected by Marvell with different $s$ respectively; Figure (c) and (d) represents the test AUC and leak AUC on the cut layer protected by Label DP with different $\epsilon$ respectively.}
 \label{fig:spectral_attack_auc_bouned_by_utility} 
 \end{figure*}

\subsection{Experimental Setup}

\textbf{Dataset.}
We mainly evaluate the proposed framework on Criteo~\footnote{{https://www.kaggle.com/c/criteo-display-ad-challenge/data}}, which's a large-scale industrial binary classification dataset (with with approximately $45$ million user click records) for conversion prediction tasks. Every record of Criteo has $27$ categorical input features and $14$ real-valued input features. We first replace all the \texttt{NA} values in categorical features with a single new category (which we represent using the empty string) and replace all the \texttt{NA} values in real-valued features with $0$. For each categorical feature, we convert each of its possible value uniquely to an integer between $0$ (inclusive) and the total number of unique categories (exclusive). For each real-valued feature, we linearly normalize it into $[0,1]$. We then randomly sample $90\%$ of the entire Criteo set as our training data and the remaining $10\%$ as our test data.  

To save space, we put the similar experimental results on another dataset Avazu in the Appendix ~\ref{sec:experimental_results_on_avazu}. 


\textbf{Model.}
 We modified a popular deep learning model architecture WDL \cite{cheng2016wide} for online advertising. Here the \nlparty first processes the categorical features in a given record by applying an embedding lookup for every categorical feature's value. We use an embedding dimension of 4 for this part. After the lookup, the deep embeddings are then concatenated with the continuous features to form the raw input vectors. The  \nlparty leverages its feature extractor $f(.)$ with $5$ layers of ReLU activated MLP to process its raw input features. The cut layer is after the output of $f(.)$. The \lparty leverages its label predictor $h(.)$ with $3$ layers of ReLU activated MLP to generate the final logit value for prediction.  
 During training, in each batch the \nlparty sends an embedding matrix with size $8,192\times 128$ to the \lparty where batch size is $8,192$\footnote{The batch size is $8,192$ if not specified.} and embedding size is $128$.

\textbf{Against Existing Defenses.} In addition to the scenario when the \lparty imposes no protection, we also include two situations when \lparty employs two existing defenses against label leakage attack in FL: Marvell~\cite{li2021label} and Label DP ~\cite{labeldp}. 

\textbf{Evaluation Metric.} 
To measure the effectiveness of the label leakage, we compute AUC on our attacker's label predictions and the ground-truth labels, namely \textit{leak AUC}. A closer to $1.0$ leak AUC is considered to be more effective, while a closer to $0.5$ leak AUC is more like a random guess.

\subsection{Results}
 
\textbf{Effectiveness of Spectral Attack.}
In this section, we show the effectiveness of leveraging Spectral attack to infer labels from the forward embedding with different settings. As shown in Figure ~\ref{fig:effectiveness_of_spectral_attack} (a), the adversarial party can infer the label information from the last two layers (including the cut layer) with a high ACU (about $0.78$) under the setting of vanilla vFL (without using any label protection techniques such as Marvell ~\cite{li2021label} and label DP~\cite{labeldp}). 
 However, as shown in Figure ~\ref{fig:effectiveness_of_spectral_attack} (b) and (c), even with the protection of  Marvell~\cite{li2021label} and Label DP~\cite{labeldp}, Spectral attack can still steal the label information with a relatively high leak AUC. Marvell~\cite{li2021label} can prevent label leakage from the backpropagated gradients by adding Gaussian noise to the gradients before sending them back to the \nlparty{}. However, as shown Figure ~\ref{fig:effectiveness_of_spectral_attack} (b), Marvell with $s=8.0$ cannot prevent label leakage from the forward embedding effectively since the last two layers can leak the label with a leak AUC closing to $0.75$. Label DP leverages randomized response to generate noisy labels for computing the loss function and the corresponding backpropagated gradients. However, as shown in Figure ~\ref{fig:effectiveness_of_spectral_attack} (c), the leak AUC of the last two layers are close to $0.75$, which indicates that Label DP fails to prevent label leakage from the forward embedding.

 \textbf{Upper Bound of Leak AUC.} As defined in our threat model in Section ~\ref{sec:threatmodel}, the adversarial party  has the freedom of selecting its network architecture (\textit{i.e.} depth and width of layers) to increase the power of $f(\cdot)$ on label guessing. One question arising is that does the power of $f(\cdot)$ on label guessing (measured by leak AUC) have an upper bound?  We may observe from Figure ~\ref{fig:effectiveness_of_spectral_attack} that the leak AUC of the cut layer is bounded by the test AUC of the model. We assume that the \lparty does not introduce negative effects on $h(\cdot)$ and hence $f(\cdot)$ at most has the same prediction power as $h(f(\cdot))$, even though the adversarial party has the freedom of increasing the depth and width of the neural network at its side to increase the power of $f(\cdot)$. 
 

 
 To illustrate this observation, we test the spectral attack AUC with different protection methods (Marvell~\cite{li2021label} and Label DP~\cite{labeldp}), since different parameters of Marvell ($s$) and Label DP ($\epsilon$) can give different model utilities. As shown in Figure ~\ref{fig:effectiveness_of_attack_on_grads} in the appendix ~\ref{sec:attack_auc_marvell_and_labeldp} and  Figure ~\ref{fig:spectral_attack_auc_bouned_by_utility}, 
 different parameters of Marvell ($s$) and Label DP ($\epsilon$) can give different trade-offs between privacy (attack AUC on backpropagated gradients as shown in Figure ~\ref{fig:effectiveness_of_attack_on_grads}) and model utilities (test AUC as shown in Figure ~\ref{fig:spectral_attack_auc_bouned_by_utility}). As shown in Figure ~\ref{fig:spectral_attack_auc_bouned_by_utility} (a) and (c), the test AUC changes with different parameters of Marvell ($s$) and Label DP ($\epsilon$). The spectral attack AUC  on the forward embedding as shown in Figure ~\ref{fig:spectral_attack_auc_bouned_by_utility} (b) and (d) has a similar pattern as the test AUC and each leak AUC is bounded by the corresponding test AUC. It indicates that the power of stealing label information from the forward embedding is bounded by the whole model's utility. 

\textbf{Attack on Balanced Datasets}. We also demonstrated the effectiveness of our attack method with balanced datasets. To save space, we put the corresponding results in Section ~\ref{sec:spectral_attack_on_balanced_datasets} of the Appendix.

%% file: tex/experiments.tex
\section{Defense Evaluation}
\label{sec:defense_evaluation}




 

 \begin{figure*}[h!]
\captionsetup[subfigure]{labelformat=empty}
\begin{subfigure}{0.245\linewidth}
  \centering
    \includegraphics[width=\linewidth]{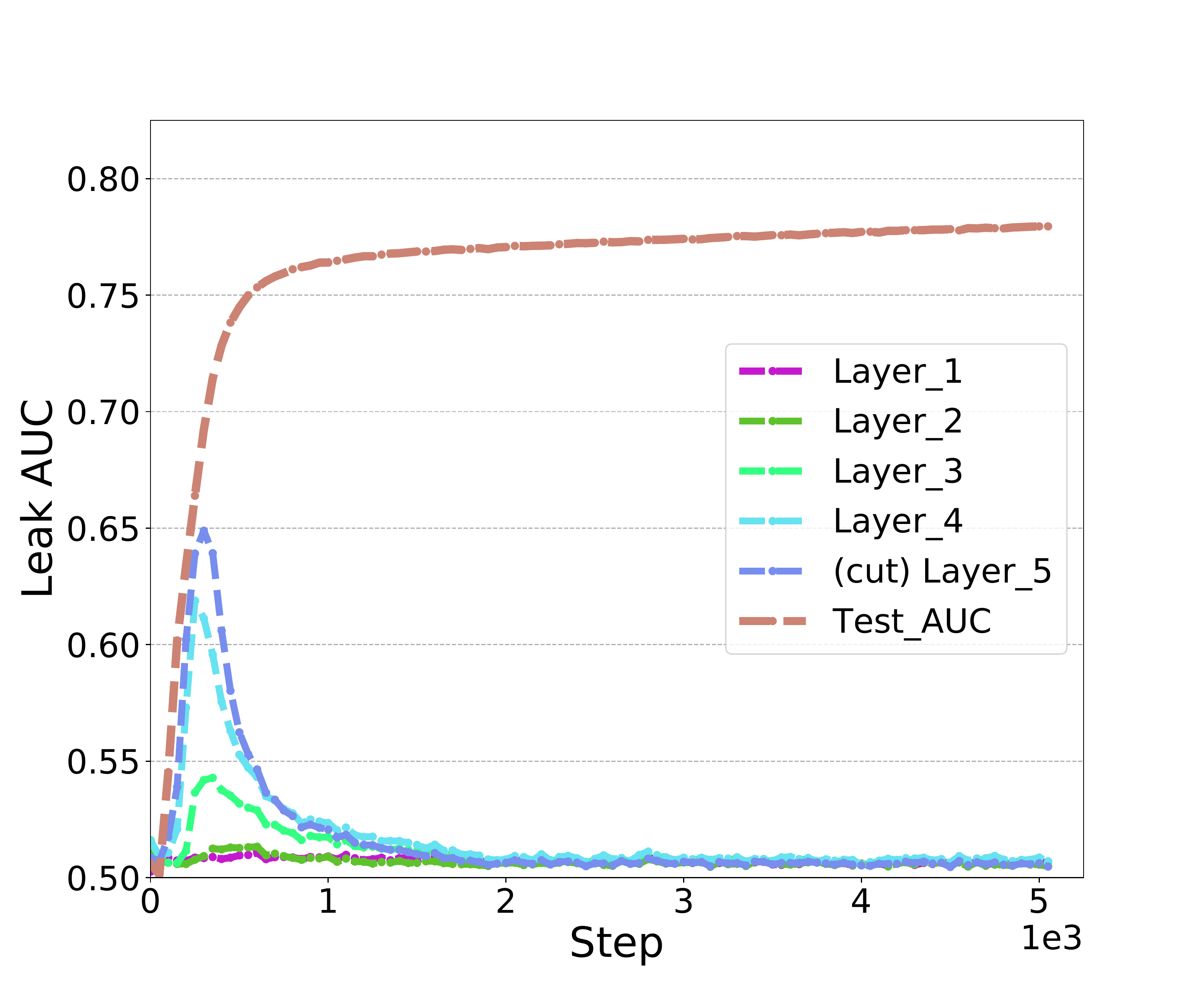}
    \caption{(a):  $\alpha_d=0.003$}
  \end{subfigure}
  \begin{subfigure}{0.245\linewidth}
  \centering
    \includegraphics[width=\linewidth]{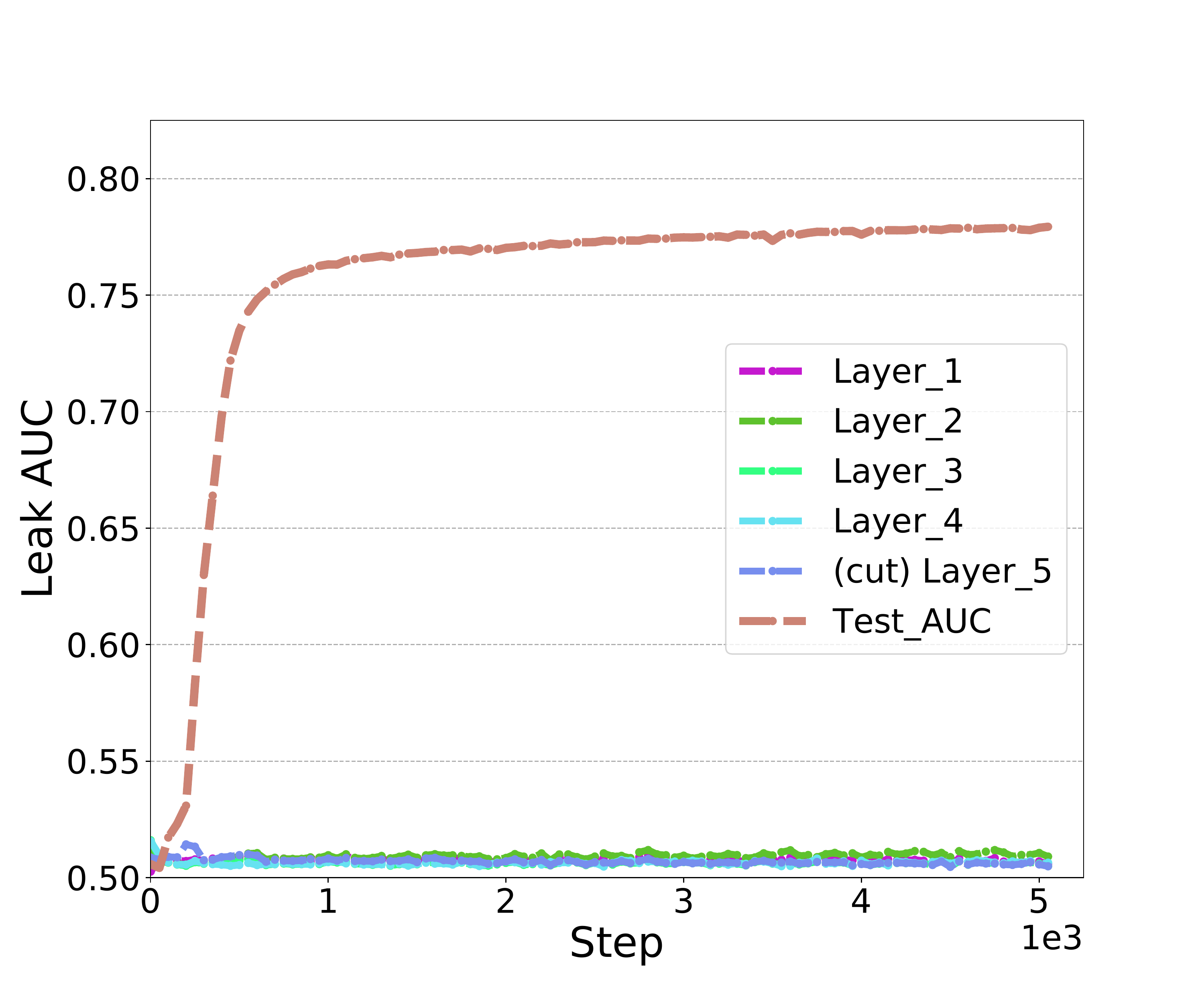}
      \caption{(b):  $\alpha_d=0.005$}
  \end{subfigure}
  \begin{subfigure}{0.245\linewidth}
  \centering
    \includegraphics[width=\linewidth]{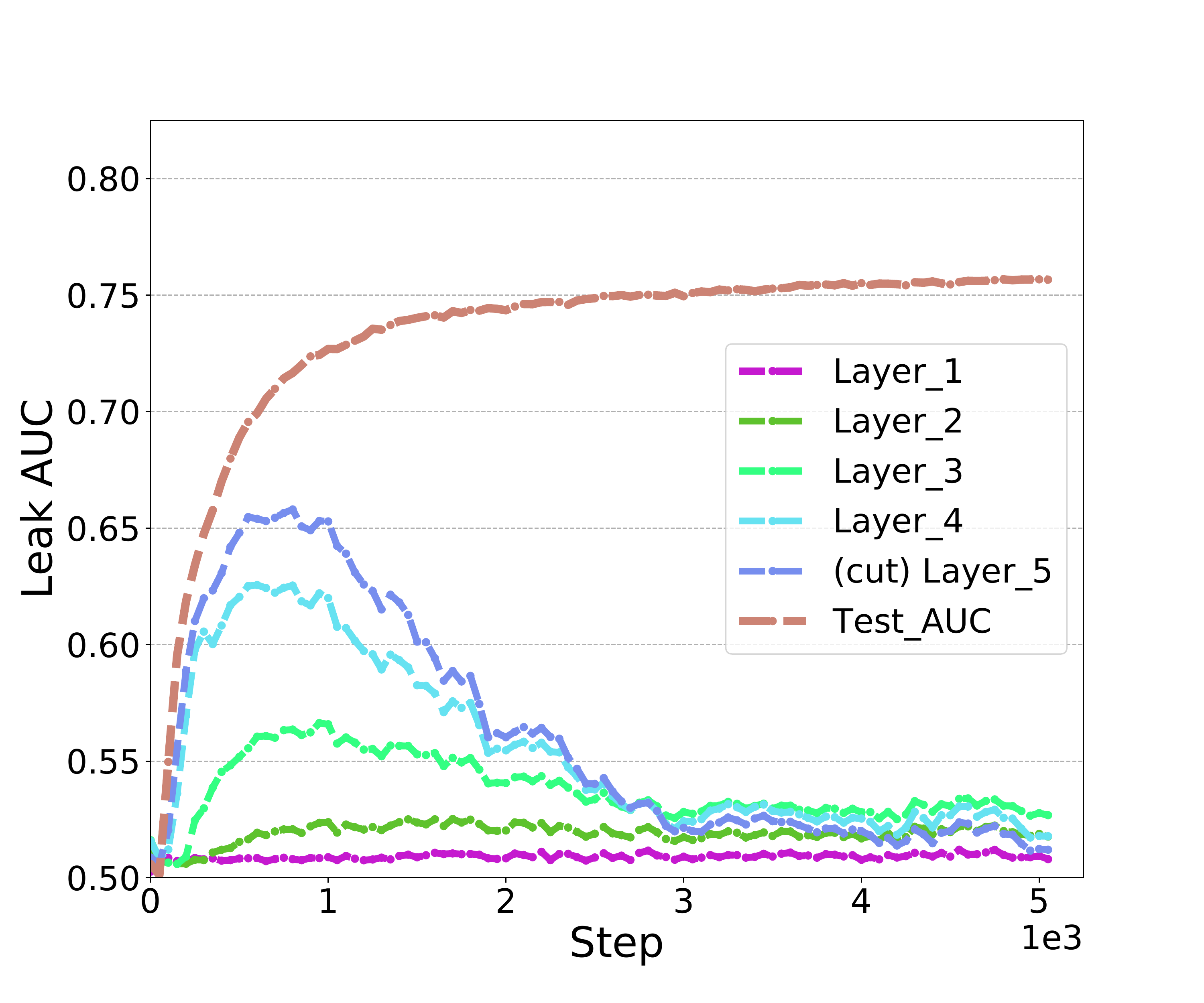}
    \caption{(c): Marvell and $\alpha_d=0.002$}
  \end{subfigure}
  \begin{subfigure}{0.245\linewidth}
  \centering
    \includegraphics[width=\linewidth]{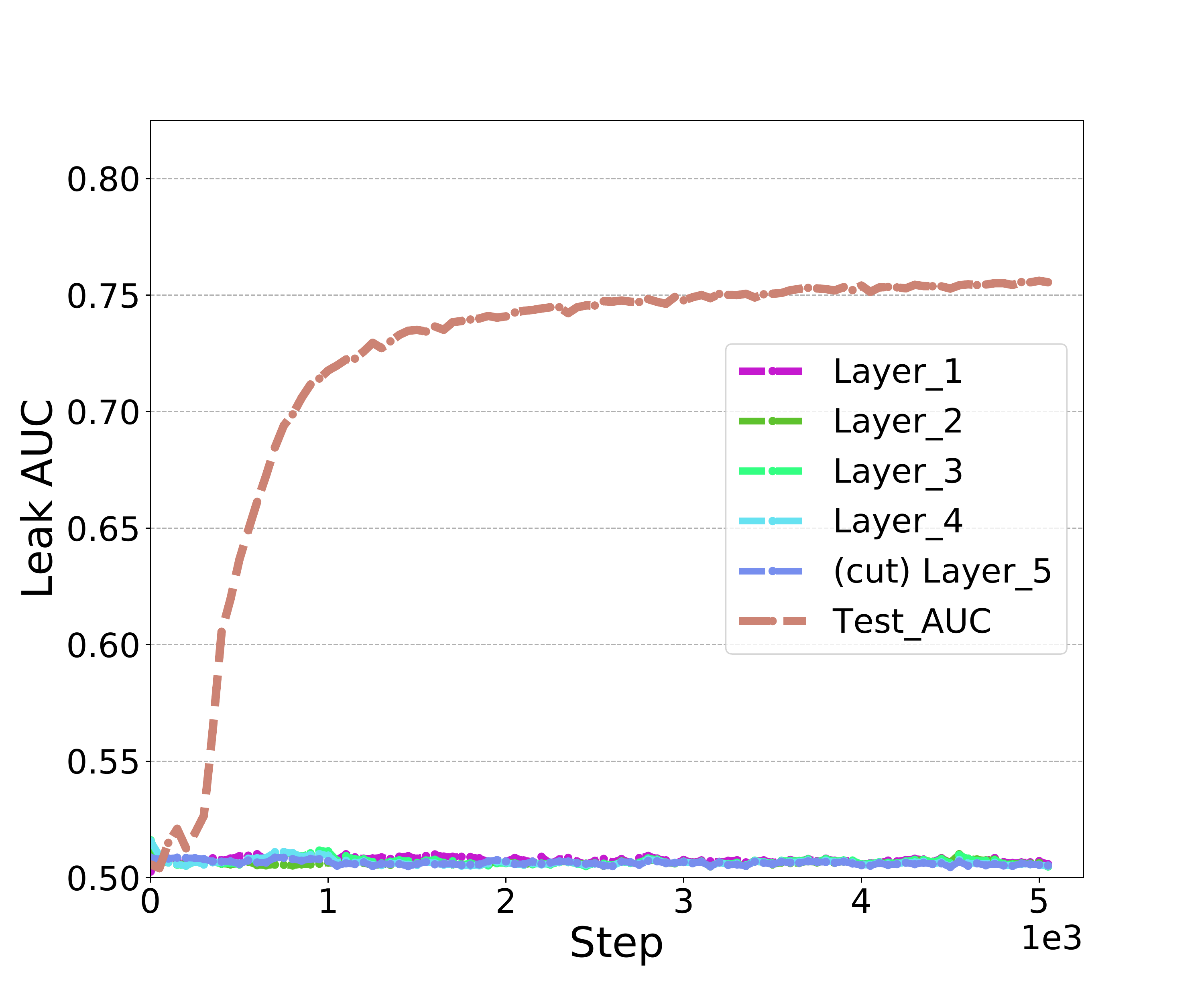}
      \caption{(d): Marvell and $\alpha_d=0.005$}
  \end{subfigure}
  \caption{Effectiveness of our protection method and its compatibility with gradient perturbation based protection method Marvell. Figure (a) and (b) shows the effectiveness of our protection method defending Spectral attack on different layers with $\alpha_d = 0.003$ and $\alpha_d = 0.005$ respectively. Figure (c) and (d) demonstrates that our protection method can be compatible with gradient perturbation based protection method Marvell $\alpha_d = 0.002$ and $\alpha_d = 0.005$ respectively.}
 \label{fig:effectiveness_of_our_protection}
 \end{figure*}
 
 \begin{figure*}[h!]
\captionsetup[subfigure]{labelformat=empty}
\begin{subfigure}{0.245\linewidth}
  \centering
    \includegraphics[width=\linewidth]{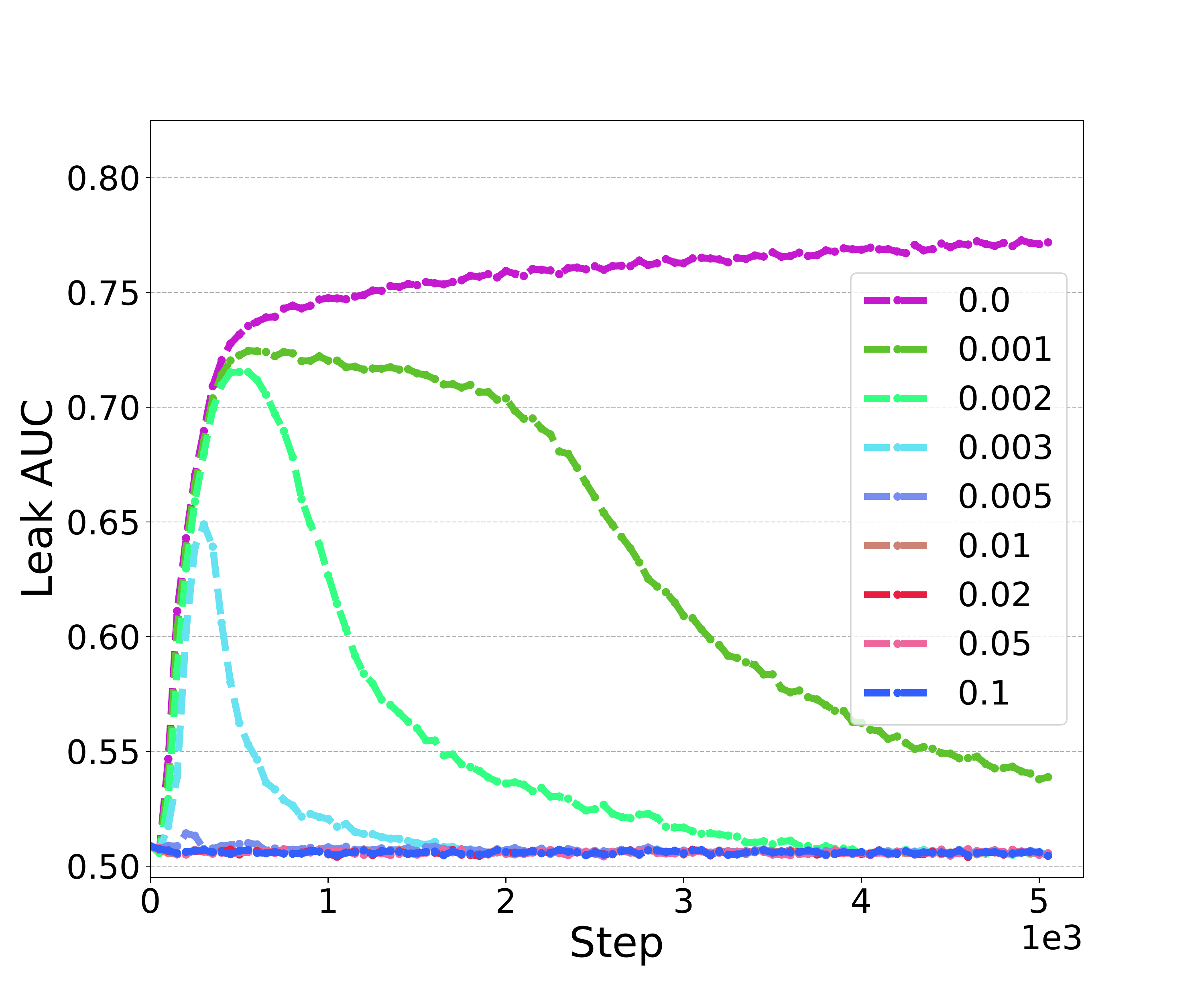}
      \caption{(a): Leak AUC (vanilla model)}
  \end{subfigure}
  \begin{subfigure}{0.245\linewidth}
  \centering
    \includegraphics[width=\linewidth]{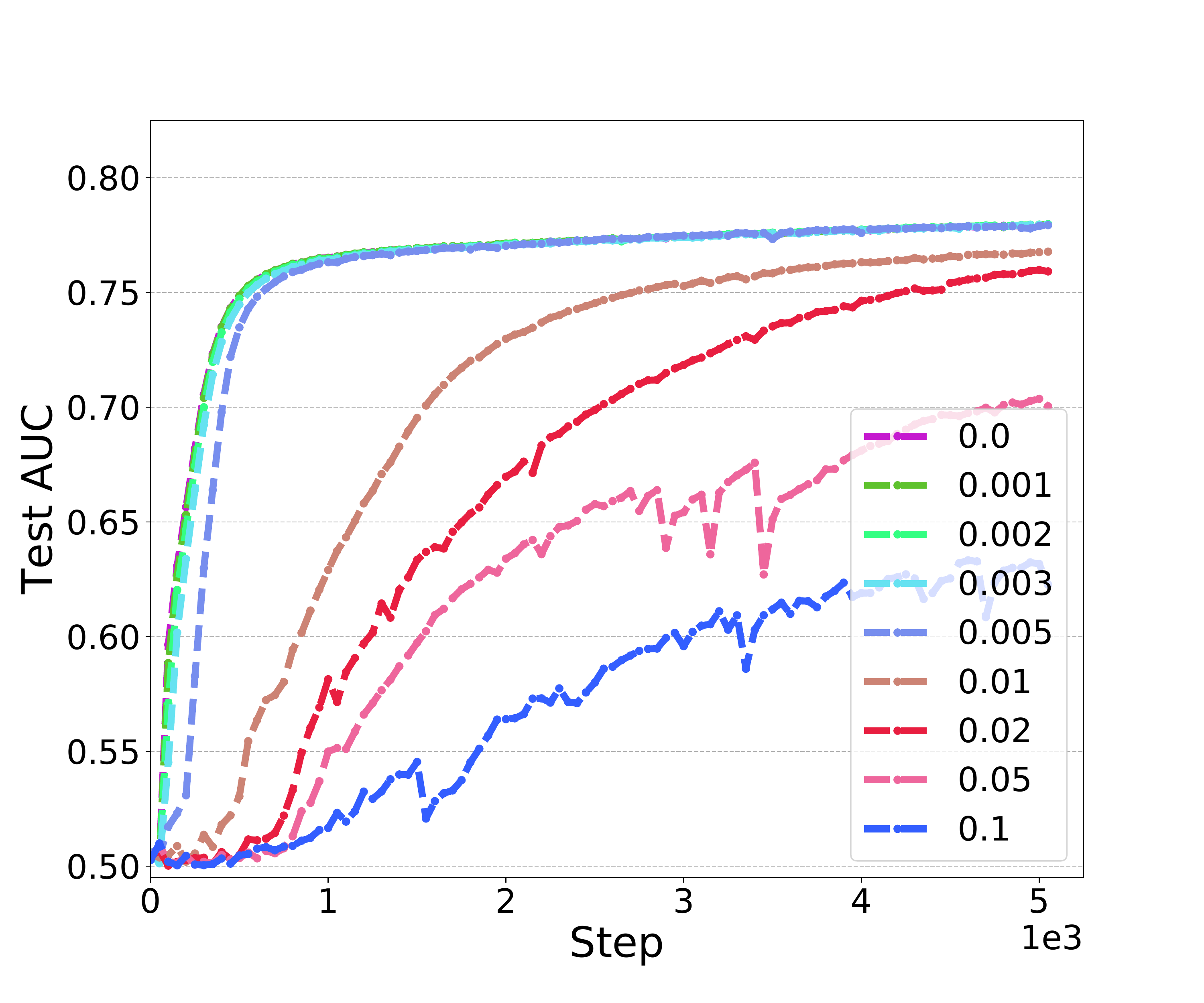}
    \caption{(b): Test AUC (vanilla model) }
  \end{subfigure}
  \begin{subfigure}{0.245\linewidth}
  \centering
    \includegraphics[width=\linewidth]{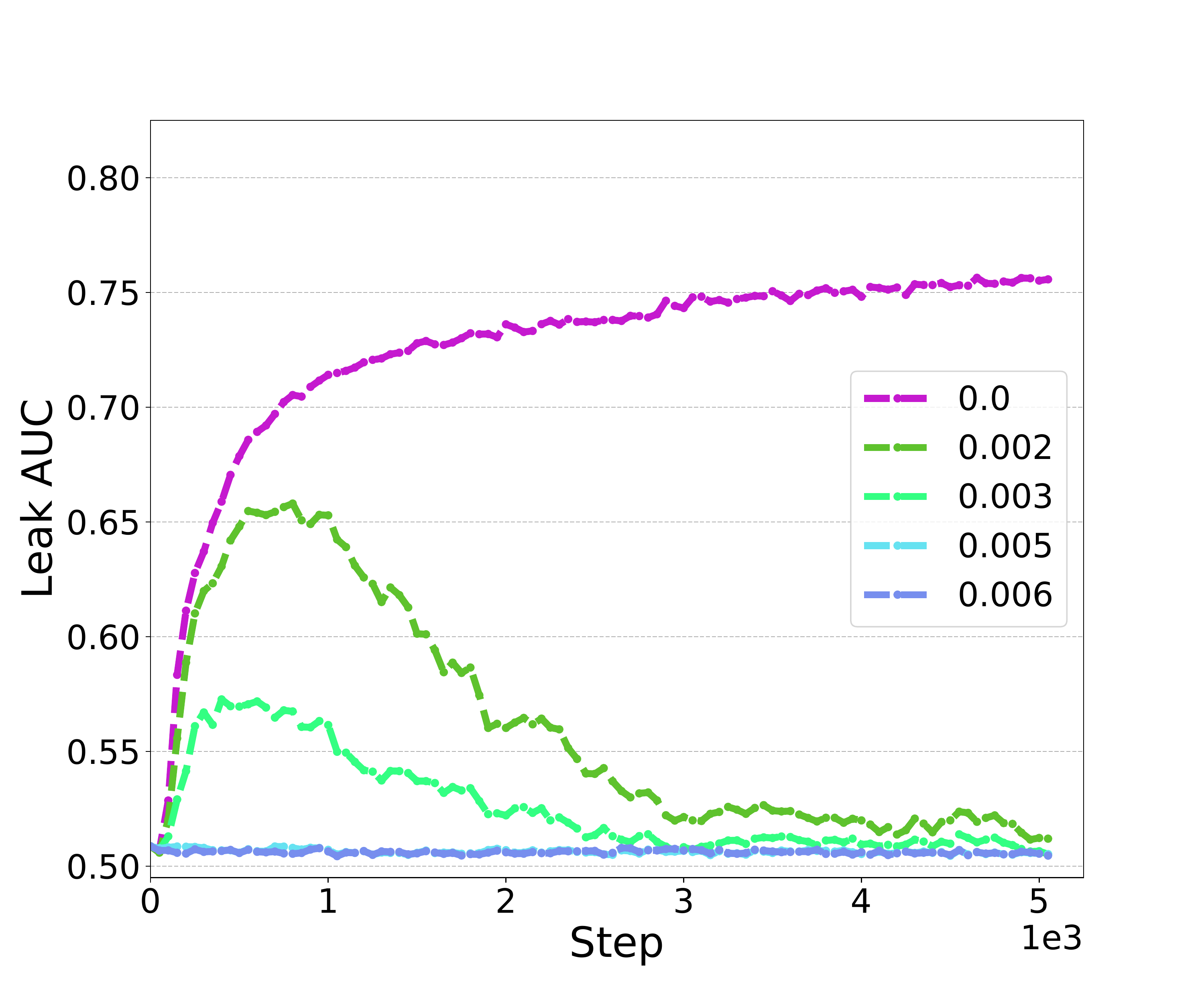}
      \caption{(c): Leak AUC (Marvell)}
  \end{subfigure}
  \begin{subfigure}{0.245\linewidth}
  \centering
    \includegraphics[width=\linewidth]{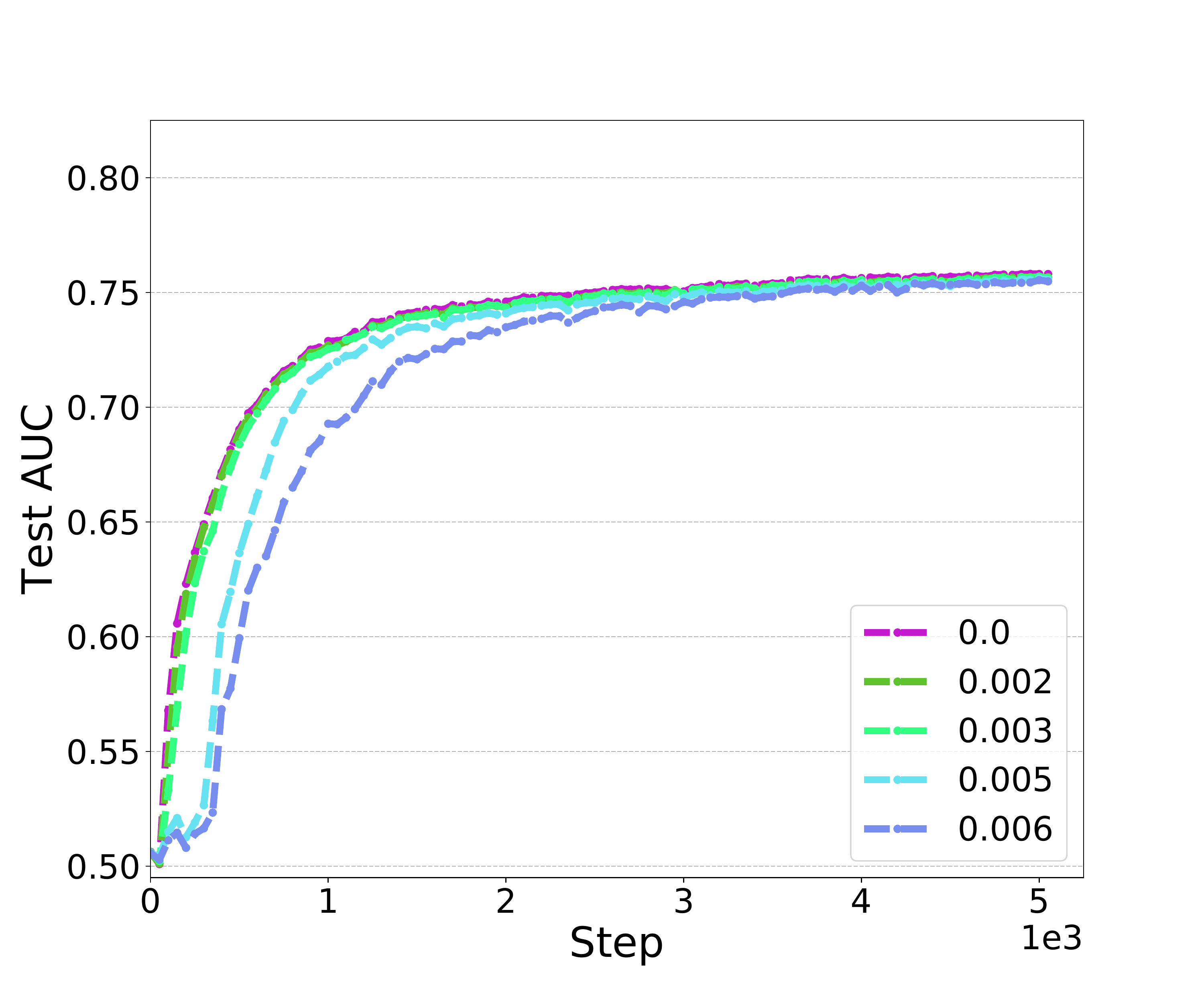}
    \caption{(e): Test AUC ( Marvell)}
  \end{subfigure}
  \caption{Sensitivity of $\alpha_d$ and tradeoff between model utility and privacy. We use different $\alpha_d$ to test the model performance and the corresponding Spectral attack AUC on the cut layer. Figure (a) and (b) shows the Spectral attack AUC and test AUC respectively. Figure (c) and (d) shows the Spectral attack AUC and test AUC with Marvell ($s=8$) respectively.}
 \label{fig:tradeoff_between_utility_privacy_varying_alpha_d} 
 \end{figure*}
 
 
 \textbf{Effectiveness.} As shown in Figure ~\ref{fig:effectiveness_of_our_protection} (a) and (b), with reasonable $\alpha_d$, the spectral attack AUC can be reduced to around $0.5$ (like a random guess), while the model utility can be as high as $0.78$ (almost the same as the performance of the vanilla model). Our protection method can be compatible with other protection methods such as Label DP~\cite{labeldp} and Marvell~\cite{li2021label}. We use Marvell as an example to demonstrate this. As shown in Figure ~\ref{fig:effectiveness_of_our_protection} (c) and (d), we can see that our method with Marvell can prevent the label leakage from the backpropagated gradients and forward embedding simultaneously.

 \textbf{Sensitivity of $\alpha_d$.}  We vary different $\alpha_d$ to test the sensitivity of $\alpha_d$ and observe the trade-off between model utility and privacy. We show model utility (test AUC) and privacy (leak AUC) with different  $\alpha_d$ in Figure ~\ref{fig:tradeoff_between_utility_privacy_varying_alpha_d}. We can conclude that a reasonable $\alpha_d$ (i.e. $0.005 \geq  \alpha_d \geq 0.002$) can achieve an acceptable attack AUC ($0.5$) while the model test AUC does not drop too much. Take $\alpha_d = 0.002$ for example, its attack AUC is around $0.5$ (as shown in Figure ~\ref{fig:tradeoff_between_utility_privacy_varying_alpha_d} (a)) and test AUC has almost no drop in comparing with the baseline ($\alpha_d = 0$) (as shown in Figure ~\ref{fig:tradeoff_between_utility_privacy_varying_alpha_d} (b)). It also happens when we use Marvell to defend the label leakage from the backpropagated gradients (as shown in Figure ~\ref{fig:tradeoff_between_utility_privacy_varying_alpha_d} (c) and (d)). It also demonstrates that our proposed protection method is compatible with these backpropagated gradients perturbation based label protection methods.

\textbf{Cost of Privacy.} Since the \lparty has minimized the distance correlation between the cut layer embedding and private labels, it has to increase its own model's power (more computational cost) to re-learn the correlation. Otherwise, it cannot maintain the whole model's utility. We provide the quantitative analysis on the cost of privacy in in the Appendix ~\ref{sec:cost_of_privacy}. 

%% file: tex/conclusion.tex
\section{Conclusion}
\label{sec:conclusion}
\vspace{-0.12in}

In this paper, we focus on identifying and defending the threat of label leakage from the forward cut layer embedding rather than backpropagated gradients. We firstly proposed a practical label attack method that can steal private labels effectively from the cut layer embedding even though some existing protection methods such as label differential privacy~\cite{labeldp} and gradients perturbation~\cite{li2021label} are applied. The effectiveness of the above label attack is inseparable from the correlation between the cut layer embedding and corresponding  private labels. Hence to mitigate the issue of label leakage from the forward embedding, we proposed to minimize the distance correlation between the cut layer embedding and corresponding private labels as an additional goal at the \lparty, which can limit the label stealing ability of the adversary. We conducted massive experiments to demonstrate the effectiveness of our proposed protection methods. Our work is the first we are aware of to identify and defend against the threat of label leakage from the forward cut layer embedding in vFL. Since the label information is highly private and sensitive, we hope our work can open up a number of worthy directions for future study in the interest of protecting label privacy.

%% file: tex/appendix.tex
\onecolumn
\textbf{Appendix Outline:}

Section ~\ref{sec:attack_auc_marvell_and_labeldp} shows the attack AUC of Marvell and Label DP with different parameters.

Section ~\ref{sec:cost_of_privacy}: Quantitative analysis on cost of privacy: shows that the \lparty has to increase its own sides' network power to achieve a good trade-off between model utility and privacy.

Section ~\ref{sec:extensions_of_protection_methods}: Extending our proposed protection methods to other settings.

Section ~\ref{sec:spectral_attack_alg}: Spectral Attack Algorithm

Section ~\ref{sec:extend_attack_to_multi_class}: Extending our attack method to multi-class scenarios

Section ~\ref{sec:experimental_results_on_avazu}: Experimental Results on Avazu

Section ~\ref{sec:spectral_attack_on_balanced_datasets}: Spectral Attack and Corresponding Defense on Balanced Datasets

Section ~\ref{sec:data_setup_experimental_details}: Data Setup and Experimental Details

\clearpage

\section{Attack AUC of Marvell and Label DP with different parameters}
\label{sec:attack_auc_marvell_and_labeldp}

\begin{figure*}[ht!]
\captionsetup[subfigure]{labelformat=empty}
\begin{subfigure}{0.245\linewidth}
  \centering
    \includegraphics[width=\linewidth]{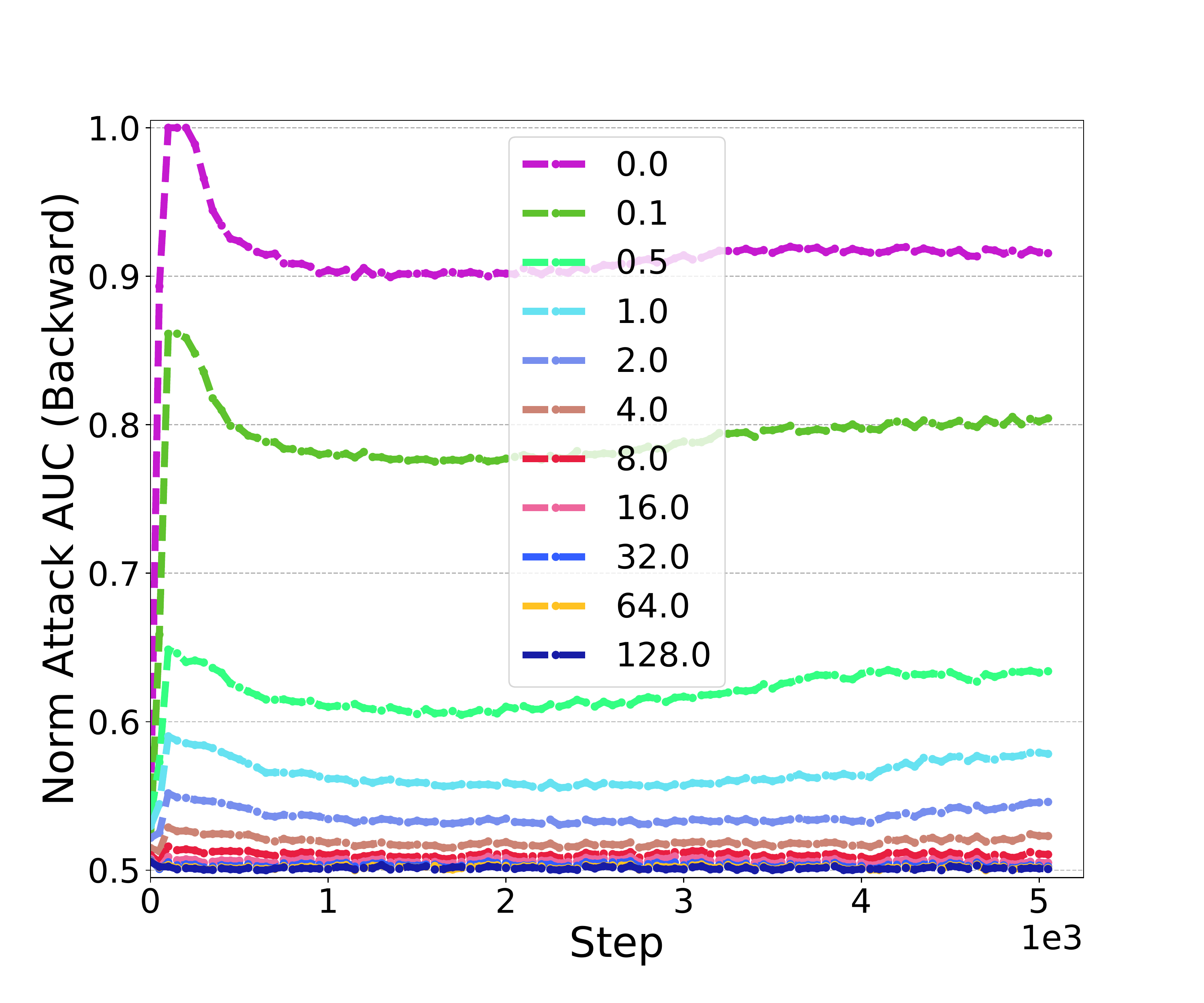}
      \caption { (a): Norm Attack, Marvell}
  \end{subfigure}
  \begin{subfigure}{0.245\linewidth}
  \centering
    \includegraphics[width=\linewidth]{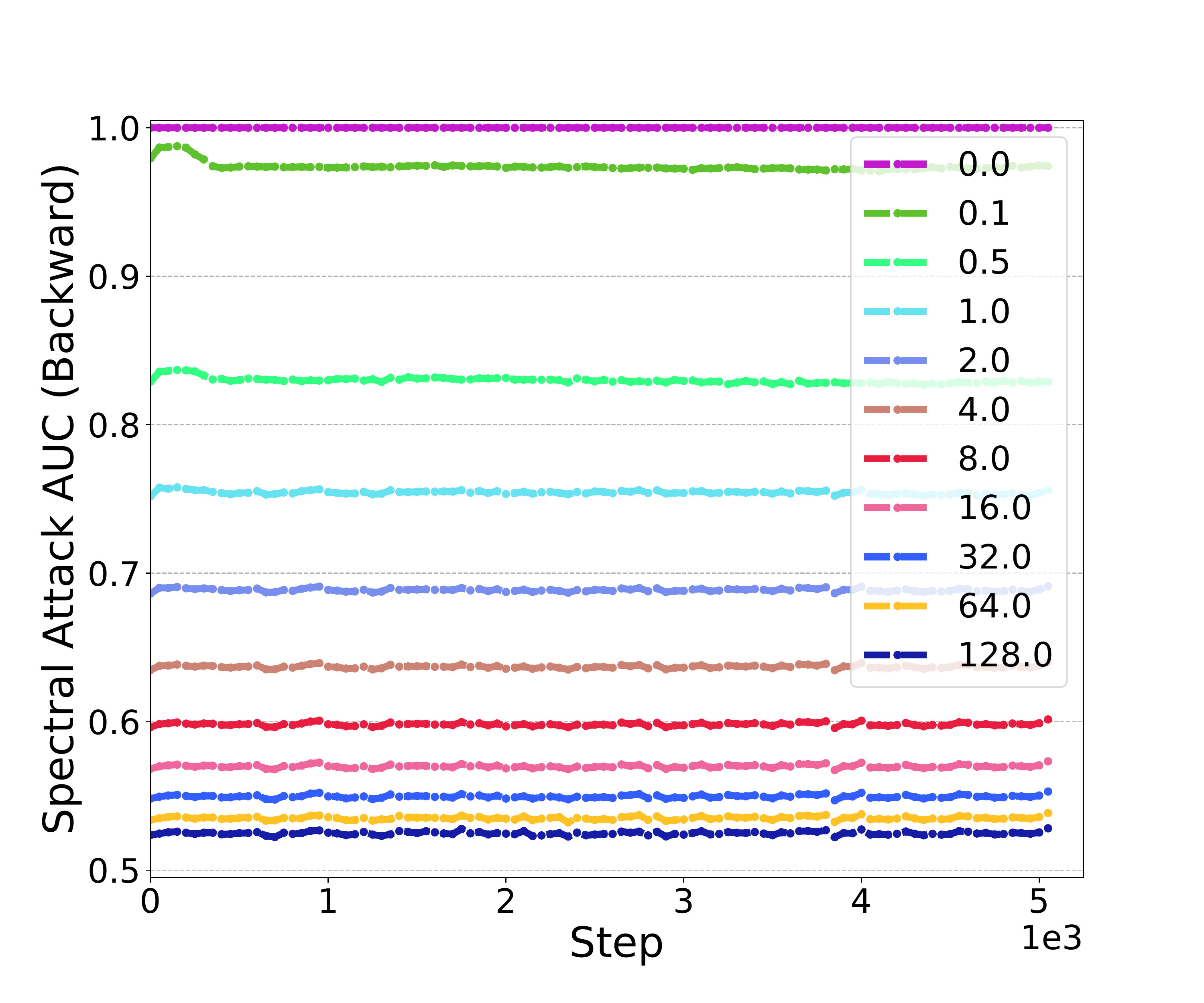}
    \caption*{(b): Spectral Attack, Marvell}
  \end{subfigure}
  \begin{subfigure}{0.245\linewidth}
  \centering
    \includegraphics[width=\linewidth]{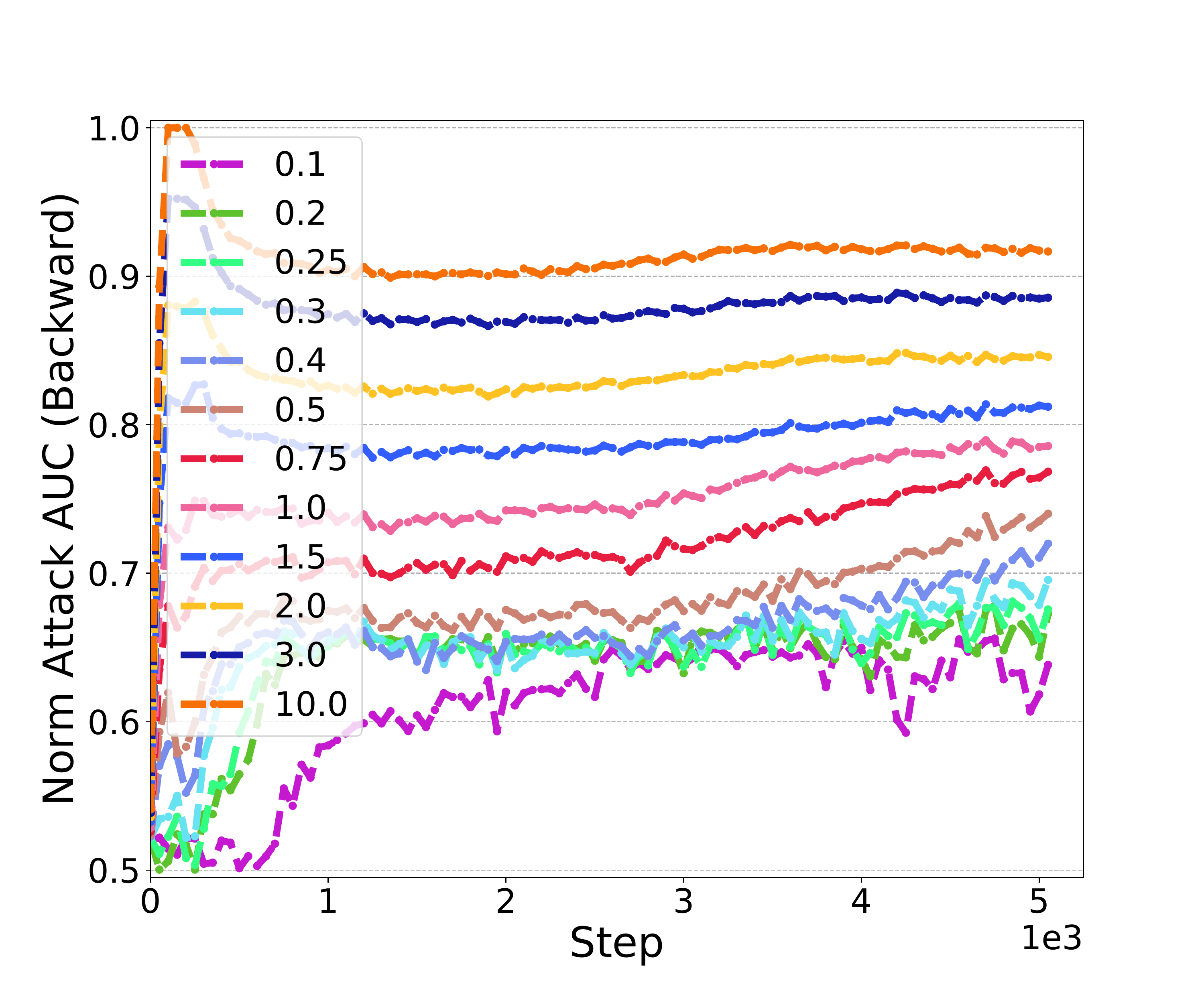}
      \caption{ (c): Norm Attack, Label DP}
  \end{subfigure}
  \begin{subfigure}{0.245\linewidth}
  \centering
    \includegraphics[width=\linewidth]{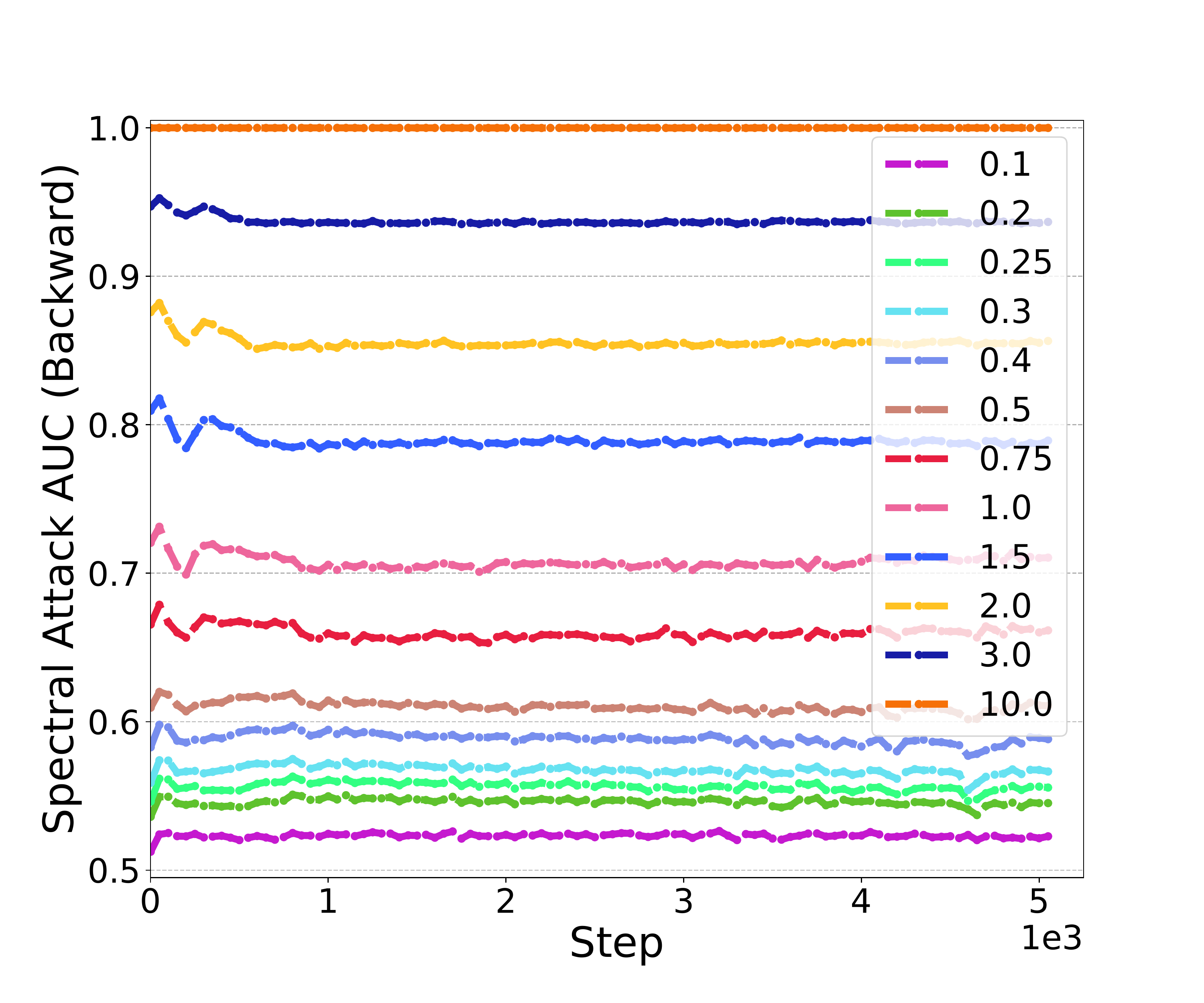}
    \caption{(b): Spectral Attack, Label DP}
  \end{subfigure}
  \caption{Effectiveness of Marvell (Figure (a) and (b)) and Label DP (Figure (c) and (d)) on defending Norm attack (Figure (a) and (c)) and Spectral attack (Figure (b) and (d)) on backpropagated (bp.) gradients. Figure (a) and (b) shows the Norm and Spectral attack AUC on the bp. gradients protected by Marvell with different $s$ respectively. Figure (c) and (d) represents the Norm and Spectral attack AUC on the bp. gradients protected by Label DP with different $\epsilon$ respectively. }
 \label{fig:effectiveness_of_attack_on_grads} 
 \end{figure*}

 \section{Quantitative Analysis on Cost of Privacy}
 \label{sec:cost_of_privacy}
 
 
 As shown in Figure ~\ref{fig:cost_of_privacy} in the Appendix ~\ref{sec:cost_of_privacy}, we report the performance of using each layer embedding to predict the label. The corresponding AUC is calculated by performing the spectral attack on each layer's embedding. The whole network consists of $8$ layers. The \nlparty owns the first $5$ layers and the \lparty owns the last $3$ layers. Without optimizing distance correlation between cut layer embedding and labels (setting $\alpha_d =0$) as shown in ~\ref{fig:cost_of_privacy} (a), (e) and (g), we can achieve comparable AUC with the final test AUC at layer $4$. It indicates that the \lparty may  need only one layer to achieve a good prediction utility. However, if we set $\alpha_d > 0$ in different settings as shown in Figure \ref{fig:cost_of_privacy} (b) ($\alpha_d = 0.002$), (c) ($\alpha_d = 0.003$), (d) ($\alpha_d = 0.005$), (f) (Marvell w. $\alpha_d = 0.002$), and (h) (Label DP w. $\alpha_d = 0.002$), the \lparty has to  use all $3$ layers to achieve a reasonable test AUC. In conclusion, the \lparty has to increase its own model's power and needs more computational cost to re-learn the correlation and achieve a reasonable trade-off between model utility and privacy.

 \begin{figure*}[h!]
\captionsetup[subfigure]{labelformat=empty}
\begin{subfigure}{0.245\linewidth}
  \centering
    \includegraphics[width=\linewidth]{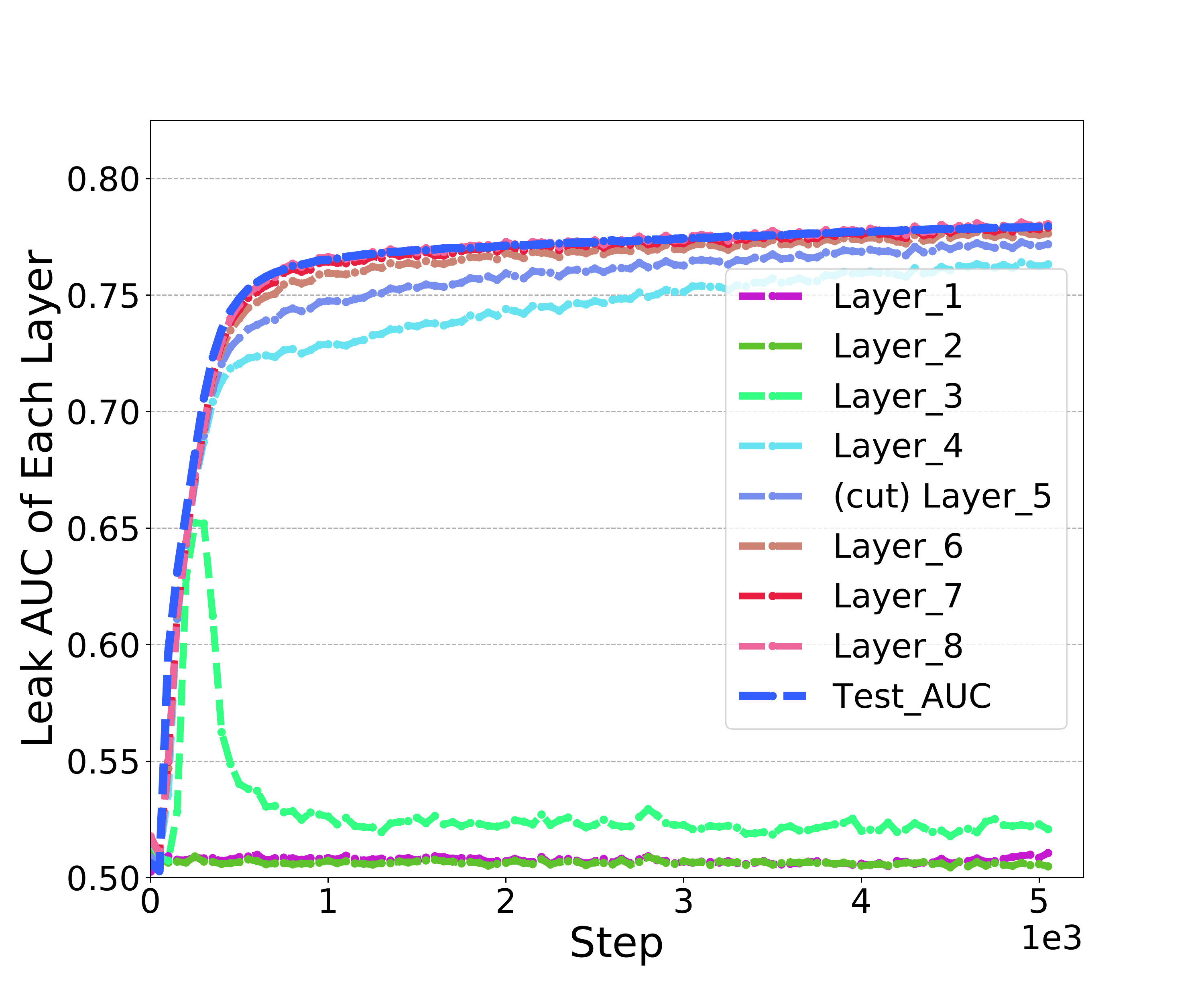} 
    \caption{(a):  vanilla model ($\alpha_d=0$)}
  \end{subfigure}
  \begin{subfigure}{0.245\linewidth}
  \centering
    \includegraphics[width=\linewidth]{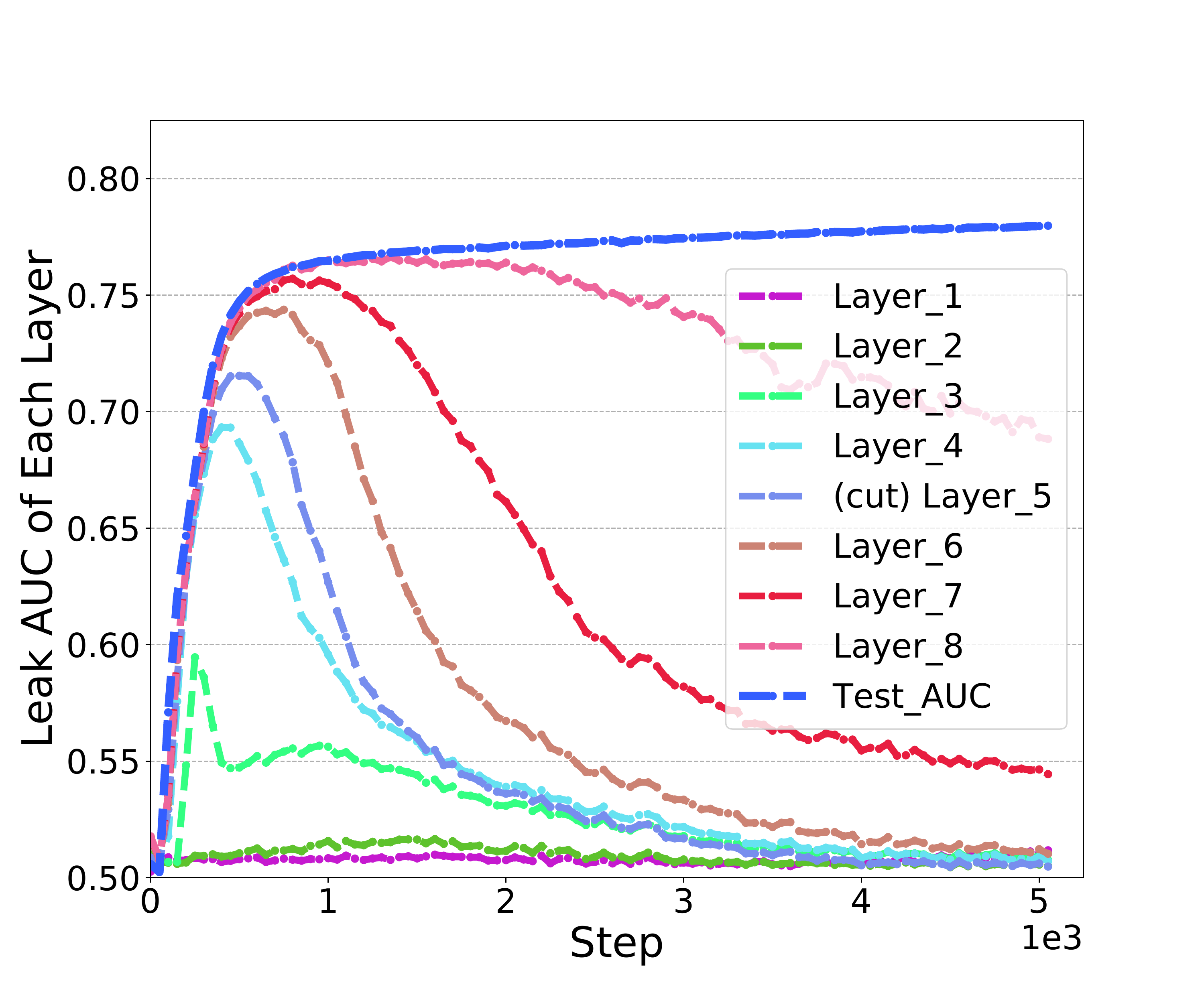}
      \caption{(b):  $\alpha_d=0.002$}
  \end{subfigure}
  \begin{subfigure}{0.245\linewidth}
  \centering
    \includegraphics[width=\linewidth]{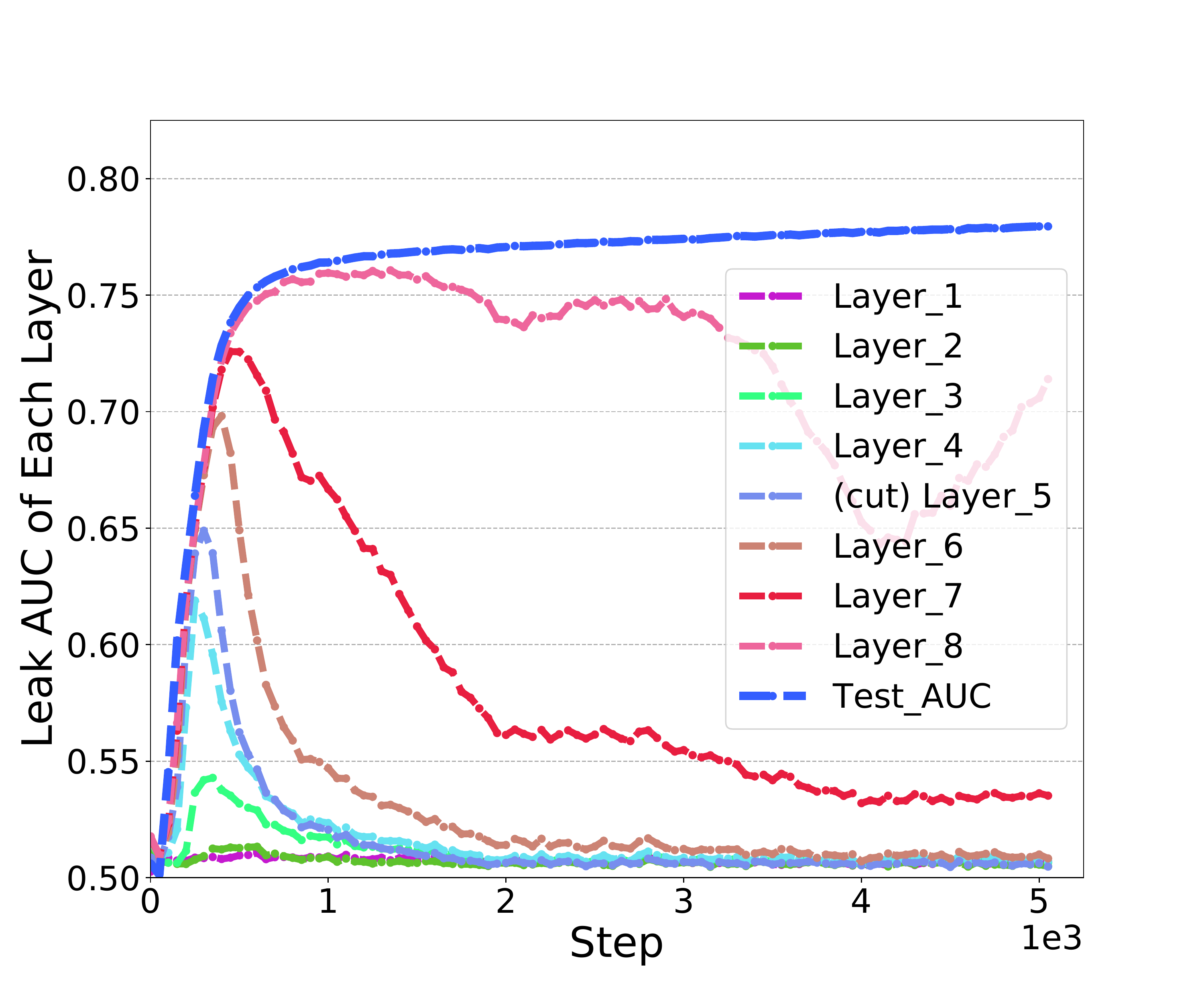}
    \caption{(c):  $\alpha_d=0.003$}
  \end{subfigure}
  \begin{subfigure}{0.245\linewidth}
  \centering
    \includegraphics[width=\linewidth]{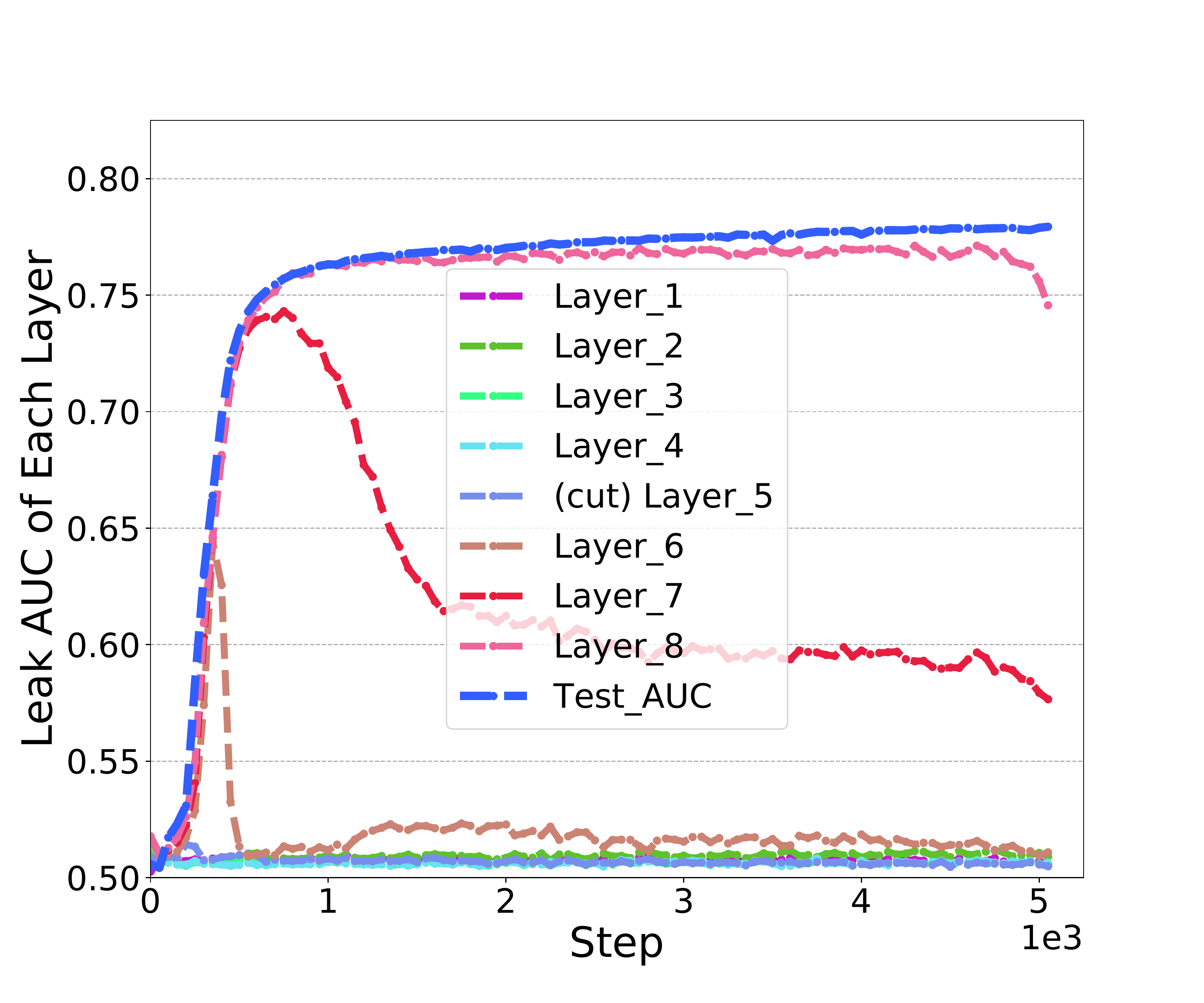}
      \caption{(d):  $\alpha_d=0.005$}
  \end{subfigure}
  
  \begin{subfigure}{0.245\linewidth}
  \centering
    \includegraphics[width=\linewidth]{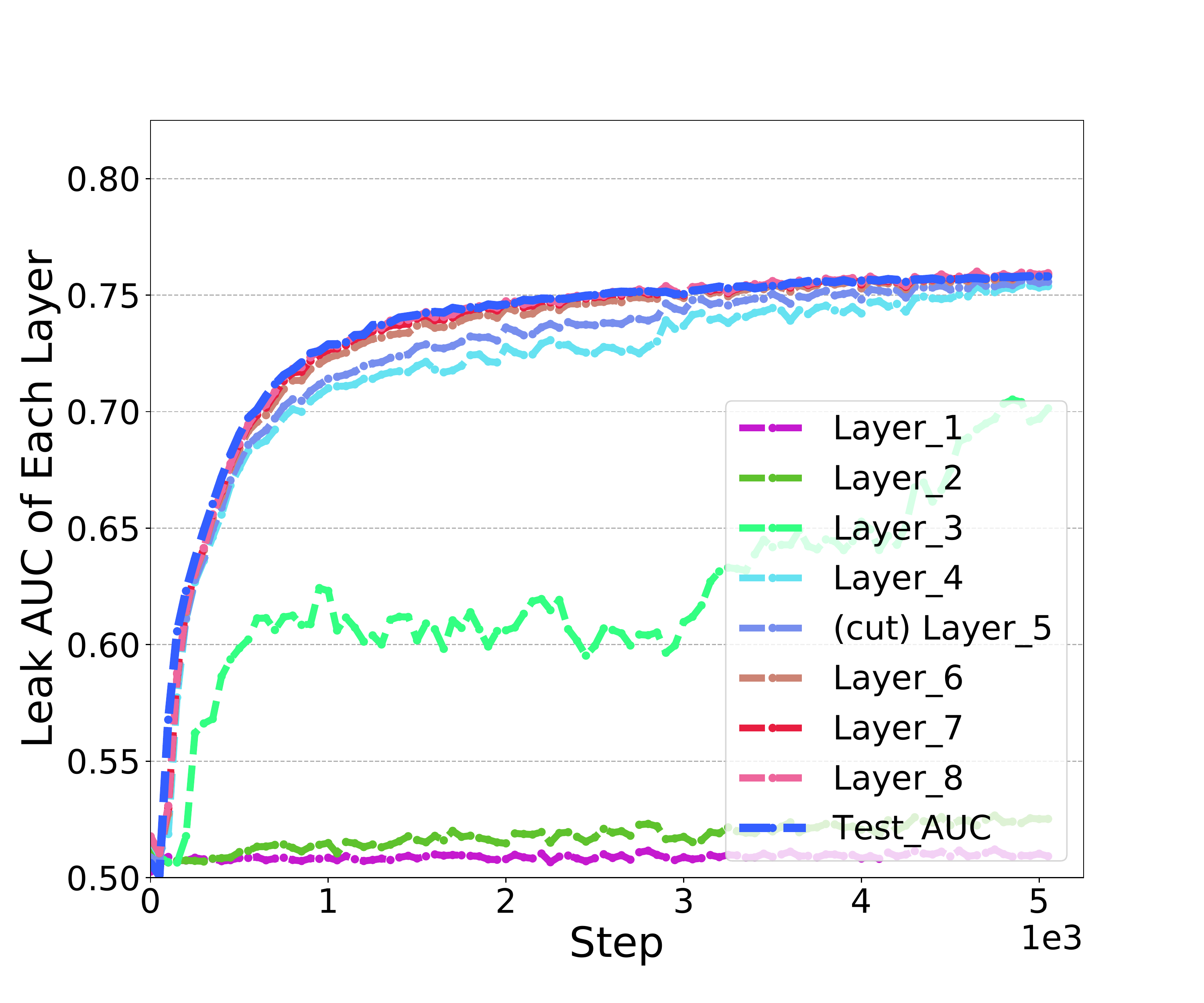}
    \caption{(e): Marvell ($s=8$) w. $\alpha_d=0$}
  \end{subfigure}
  \begin{subfigure}{0.245\linewidth}
  \centering
    \includegraphics[width=\linewidth]{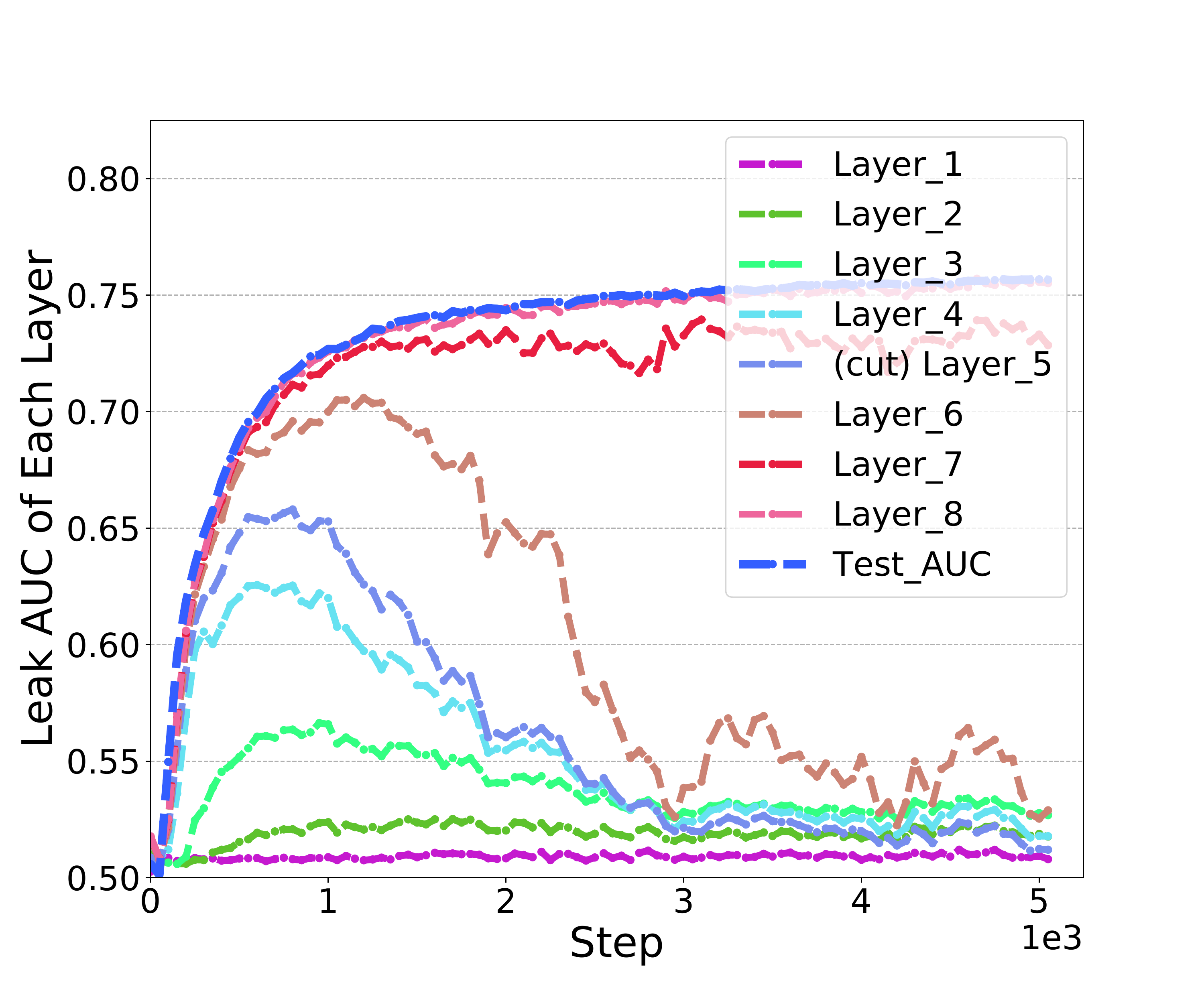}
      \caption{(f): Marvell ($s=8$) w. $\alpha_d=0.002$}
  \end{subfigure}
  \begin{subfigure}{0.245\linewidth}
  \centering
    \includegraphics[width=\linewidth]{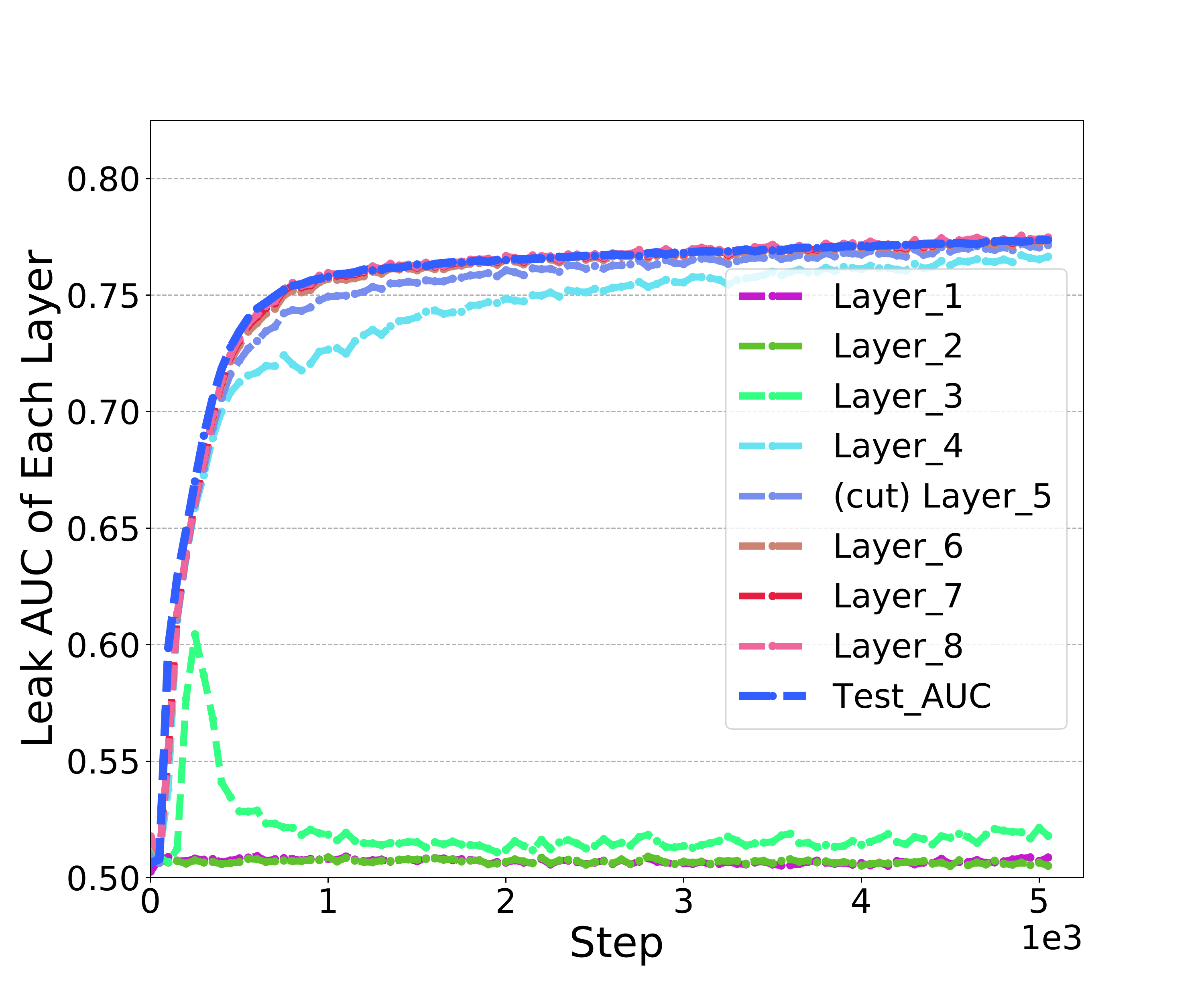}
    \caption{(g): Label DP  ($\epsilon=2$)  $\alpha_d=0$}
  \end{subfigure}
  \begin{subfigure}{0.245\linewidth}
  \centering
    \includegraphics[width=\linewidth]{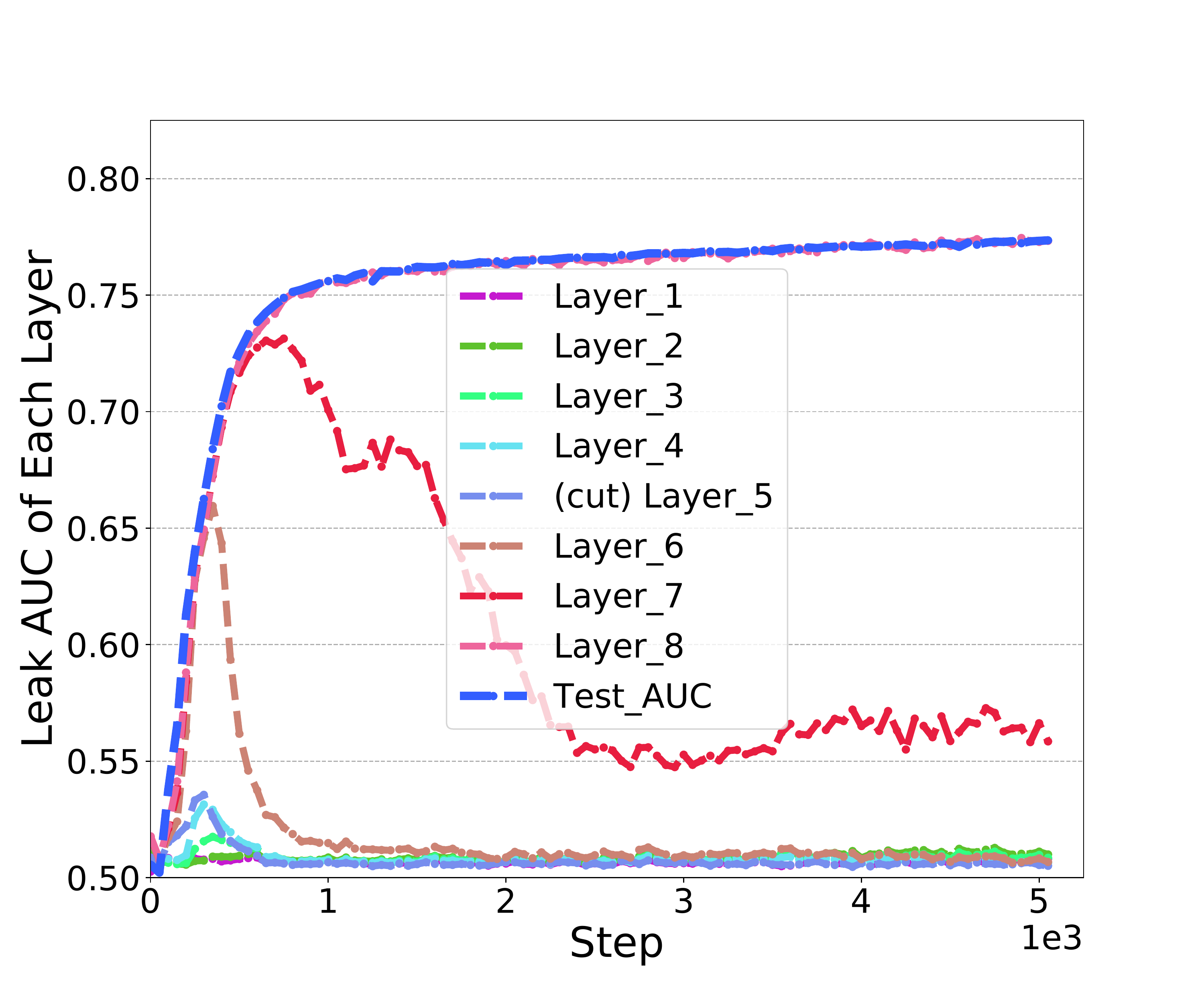}
      \caption{(h): Label DP ($\epsilon=2$) w. $\alpha_d=0.002$}
  \end{subfigure}
  \caption{Cost of privacy. Since the \lparty has minimized the distance correlation between the cut layer embedding and label, it has to increase its own neural network's power to relearn the correlation and maintain the whole model's utility. We report the leak AUC of using each layer embedding to predict the label. The corresponding AUC is calculated by performing spectral attack on each layer's embedding. The whole network consists of 8 layers. The \nlparty owns the first $5$ layers and the \lparty owns the last $3$ layers. Without optimizing distance correlation between cut layer embedding and labels ($\alpha_d =0$) as shown in Figure (a), (e) and (g), we can achieve comparable AUC with the final test AUC at the forth and its following layers. It indicates that the \lparty may not need $3$ layers to achieve a good prediction utility. However, if we set $\alpha_d > 0$ in different settings (w/wo. Label DP and w/wo.  Marvell), the \lparty has to fully used its $3$ layers to achieve a reasonable test AUC. In conclusion, the \lparty has to increase its own model's power to achieve a good trade-off between privacy and model utility. }
 \label{fig:cost_of_privacy}
 \end{figure*}
 
To provide more quantitative analysis of privacy cost, we conducted a new round of experiments to analyze the cost of privacy on Criteo dataset as shown in Table ~\ref{tab:cost_of_privacy_analysis}. We change the predictive power of the label party by varying the number of layers in the label party's neural network. For each $\alpha_d$, the label party trains a MLP model with 1 layer (small model) and 3 layers (large model) respectively and we report the corresponding test AUCs. As a result, when we increase $\alpha_d$, the test AUCs of both small and large models decrease. (The same conclusion is shown in Figure ~\ref{fig:tradeoff_between_utility_privacy_varying_alpha_d} (a) and (b).) This is a natural trade-off between utility and privacy.
 We also observe that when $\alpha_d$ is not large (i.e. $\leq 0.005$), the gap of test AUC between small and large model is not significant. It indicates that it would be easy for the label party to re-learn the correlation between cut layers embedding and label information with only 1 layer on small $\alpha_d$. However when increasing $\alpha_d$, the difference of test AUC of small and large models becomes larger. For example, when $\alpha_d = 0.03$, if the label party has only 1 layer, its test AUC is only $0.7082$. Meanwhile if the label party increases its model's predictive power by expanding its model to 3 layers, the corresponding test AUC can be increased to $0.7518$. With $\alpha_d=0.02$, the label party can only use 1 layer to reach the similar level of test AUC ($0.7528$). Hence the label party has to use more layers (and more computational cost) to increase its model's predictive power to re-learn the correlation between cut layer embedding and label information for larger $\alpha_d$. 
 
\begin{table}[!htp]\centering
\begin{tabular}{cc|cc}
& $\alpha_d$ & Test AUC (small) & Test AUC (large) \\ \hline 
&0.003 &0.7769 &0.7777 \\
&0.005 &0.7763 &0.7773 \\
&0.02 &0.7528 &0.7645 \\
&0.03 &0.7082 &0.7518 \\
&0.04 &0.5848 &0.7152 \\
\bottomrule
\end{tabular}
\caption{Test AUC on Criteo with varying $\alpha_d$ and model's predictive power (small 1-layer or large 3-layer model).}\label{tab:cost_of_privacy_analysis}
\end{table}

\section{Extending Our Protection Methods to Other Settings}
\label{sec:extensions_of_protection_methods}

We choose the two-party setting with binary classification because it is currently the typical scenario used in the vFL literature (i.e. \cite{lzw20,li2021label}) due to its popularity in the industry (online advertising, disease prediction etc). 
Our protection method can be easily extended to other  scenarios as following:

\begin{itemize}
\setlength\itemsep{0pt}
    \item {Multi-party vFL}: Each \lparty can add an additional loss function $\mathcal{L}_{d}^i$ to minimize the distance correlation between the private label and the  forward embedding sent from the \nlparty{} $i$. The rest of the training would be the same.
    \item {Multi-class dataset}: The calculation of distance correlation between the forward embedding and private labels $Y$ does not require $Y$ to be binary.
    \item {Label party also owns features}: the \lparty does not have to limit the correlation between its own features and its own labels. The training would be the same. 
\end{itemize}

\section{Spectral Attack Algorithm}
\label{sec:spectral_attack_alg}

The detailed description of  our attack method can be seen in Algorithm ~\ref{alg:spectral_attck}.

\begin{algorithm}[H]
\DontPrintSemicolon
\SetAlgoLined
\KwResult{Infer Labels for $n$ samples in a mini-batch}
\SetKwInOut{Input}{Input}\SetKwInOut{Output}{Output}
\Input{Cut layer embedding $f(X) \in \mathcal{R}^{n \times d}$}
\Output{Inferred labels of $f(X)$}
\BlankLine

$\text{mean}_f = mean(f(X), axis = 0)$ \quad // output dim: $1 \times d$

$\text{normalized}_f = f(X) - \text{mean}_f$ \quad // output dim: $n \times d$

// S, U, and V are singular values, right singular vectors and left singular vectors of $normalized_f$ respectively

S, U, V = svd($normalized_f$) \quad 

$v = V^T[0]$ \quad // output dim: $1 \times d$

$s = \langle \text{normalized}_f, v \rangle$ \quad // output dim: $n \times 1$

Cluster $s$ into two clusters: $C_1$ and $C_2$

\eIf{$|C_1| > |C_2|$}{
        assign negative labels to samples in $C_1$\; 
        assign positive labels to samples in $C_2$\;
    }{
         assign negative labels to samples in $C_2$\; 
        assign positive labels to samples in $C_1$\;
    }
\caption{Spectral Attack Algorithm}
\label{alg:spectral_attck}
\end{algorithm}

\section{Extending Our Attack Method to Multi-class Scenarios}
\label{sec:extend_attack_to_multi_class}

Our attack method can be easily extended to a multi-class task. Suppose the dataset has k classes, we can use multi-class culstering algorithms such as k-means clustering and spectral clustering on the embeddings to generate $k$ clusters. Then we can assign the attacker's inferring labels to k clusters with the global knowledge on the label ratios. For example, the label of the largest class (class with the most instances) is assigned to the largest cluster, and so on.

\section{Experimental Results on Avazu Dataset}
\label{sec:experimental_results_on_avazu}

We also conducted experiments on Avazu dataset, which is an online advertising dataset containing approximately $40$ million entries ($11$ days of clicks/not clicks). The positive ratio is about $15\%$. We randomly split the dataset into $90\%$ training and $10\%$ test, and run $10$ $epochs$ with $8,192$ batch size to train the model. Other settings such as the network structure are the same as the first Criteo dataset in the paper. 

We show the results in Table ~\ref{tab:results_on_avazu}. When there is no protection ($\alpha_d=0$), the Leak AUC is $0.7262$ which is much higher than the random guess attack ($0.5$). When we leverage our protection method ($\alpha_d =0.03$), the Leak AUC drops to $0.5$. Meanwhile, the test AUC of $\alpha_d =0.03$ only drops about $0.003$ when comparing with the model without protection. In general, the overall observations and conclusions on Avazu are consistent with Criteo results in the paper. 

\begin{table}[!htp]\centering
\begin{tabular}{c|c|c}\toprule
&Leak AUC &Test AUC \\ \hline 
vanilla ($\alpha_d=0$) &0.7262 &0.7532 \\ \hline 
$\alpha_d=0.03$ &0.5089 &0.7502 \\
\bottomrule
\end{tabular}
\caption{Results on Avazu Dataset}\label{tab:results_on_avazu}
\end{table}

\section{Spectral Attack on Balanced Datasets}
\label{sec:spectral_attack_on_balanced_datasets}

In terms of the balanced setting, we artificially down-sample the negative instances in Criteo, and generate three datasets with positive ratio $50\%$, $49\%$ and $45\%$. Then we conduct a similar analysis on those datasets, and show the results in Table ~\ref{tab:criteo_pos_ratio_0.5}, ~\ref{tab:criteo_pos_ratio_0.49}, and Table ~\ref{tab:criteo_pos_ratio_0.45}.

Note that when the dataset is balanced (i.e. positive ratio is $0.5$), it is true that the attacker cannot directly know which label to assign to which cluster. However, the attacker can employ a simple heuristic: assigning positive labels to the cluster with larger scores ($s$ calculated in Step 1 in Section ~\ref{sec:label_leakage_problem}). As shown in Table ~\ref{tab:criteo_pos_ratio_0.5}, the attack can still be highly successful ($0.7817$ leak AUC without protection). In addition, our defense in the balanced setting can also successfully bring down the leak AUC to $0.5$. We observe the similar conclusions when the positive ratio is $49\%$ and $45\%$. Those results show that our defense can work when the dataset is balanced. 

\begin{table}[!htp]\centering
\begin{tabular}{c|c|c}\toprule
 &Leak AUC &Test AUC \\ \hline
$\alpha_d = 0$ &0.7607 &0.7823 \\ \hline
$\alpha_d = 0.03$  &0.5048 &0.7824 \\
\bottomrule
\end{tabular}
\caption{Results on Criteo with Positive Ratio as  $45\%$.}\label{tab:criteo_pos_ratio_0.45}
\end{table}

\begin{table}[!htp]\centering
\begin{tabular}{c|c|c}\toprule
 &Leak AUC &Test AUC \\ \hline
$\alpha_d = 0$  &0.7814 &0.7849 \\ \hline
$\alpha_d = 0.03$  &0.5069 &0.7837 \\
\bottomrule
\end{tabular}
\caption{Results on Criteo with Positive Ratio as  $49\%$.}\label{tab:criteo_pos_ratio_0.49}
\end{table}

\begin{table}[!htp]\centering
\begin{tabular}{c|c|c}\toprule
&Leak AUC &Test AUC \\ \hline
$\alpha_d = 0$ &0.7817 &0.7829 \\ \hline
$\alpha_d = 0.03$ &0.5061 &0.7826 \\
\bottomrule
\end{tabular}
\caption{Results on Criteo with Positive Ratio as  $50\%$.}\label{tab:criteo_pos_ratio_0.5}
\end{table}

\section{Data Setup and Experimental Details}
\label{sec:data_setup_experimental_details}

We first describe how we first preprocess each of the datasets. We then describe the model architecture used for each dataset. Finally, we describe what are the training hyperparameters used for each dataset/model combination and the total amount of compute required for the experiments.

\subsection*{Dataset preprocessing}

\paragraph{[Criteo]} Every record of Criteo has $27$ categorical input features and $14$ real-valued input features. We first replace all the \texttt{NA} values in categorical features with a single new category (which we represent using the empty string) and all the \texttt{NA} values in real-valued features by $0$. For each categorical feature, we convert each of its possible value uniquely to an integer between $0$ (inclusive) and the total number of unique categories (exclusive). For each real-valued feature, we linearly normalize it into $[0,1]$. We then randomly sample $10\%$ of the entire Criteo publicly provided training set as our entire dataset (for faster training to generate privacy-utility trade-off comparision) and further make the subsampled dataset into a 90\%-10\% train-test split.

\paragraph{[Avazu]} Unlike Criteo, each record in Avazu only has categorical input features. We similarly replace all \texttt{NA} value with a single new category (the empty string), and for each categorical feature, we convert each of its possible value uniquely to an integer between $0$ (inclusive) and the total number of unique categories (exclusive). We use all the records in provided in Avazu and randomly split it into 90\% for training and 10\% for test.

\subsection*{Model architecture details}

\paragraph{[Criteo]} We modified a popular deep learning model architecture WDL \cite{cheng2016wide} for online advertising. We first process the categorical features in a given record by applying an embedding lookup for every categorical feature's value. We use an embedding dimension of 4 for the deep part. After the lookup, the deep embeddings are then concatenated with the continuous features to form the raw input vectors for the deep part. The deep part processes the raw features using several ReLU-activated 128-unit MLP layers before producing a final logic value. 

\subsection*{ Model training details}

To ensure smooth optimization and sufficient training loss minimization, we use a slightly smaller learning rate than normal.

\paragraph{[Criteo]}
We use the Adam optimizer with with a batch size of 8,192 and a learning rate of $1e$$-4$ throughout the entire training of 3 epochs (approximately 15k stochastic gradient updates).

\paragraph{[Avazu]}
We use the Adam optimizer with a batch size of 8,192 and a learning rate of $1e$$-4$ throughout the entire training of 3 epochs (approximately 15k stochastic gradient updates).

We conduct our experiments over 8 Nvidia Tesla V100 GPU card. Each epoch of run of Avazu takes about 10 hours to finish on a single GPU card occupying 4GB of GPU RAM. Each epoch run of Criteo takes about 11 hours to finish on a single GPU card using 4 GB of GPU RAM.